\documentclass[opre,nonblindrev]{informs3}
\usepackage{changes}
\usepackage{longtable}
\usepackage{pifont}
\usepackage{kbordermatrix}
\usepackage{algorithm,float}
\usepackage[noend]{algpseudocode}
\usepackage{booktabs}
\usepackage{longtable}
\usepackage{pifont}
\usepackage{subfigure}
\usepackage{graphicx}
\usepackage{makecell}
\setcellgapes{4pt}
\allowdisplaybreaks
\newcommand{\pr}{\mathbb{P}} 
\newcommand{\E}{\mathbb{E}} 
\newcommand{\dR}{\mathbb{R}} 
\newcommand\ehist[1]{\ensuremath{{#1}, \tilde M_{#1}}}
\newcommand\hist[1]{\ensuremath{[{#1}, \tilde M_{#1}]}}

\newcommand\ehisttt[1]{\ensuremath{{#1}^{t-1}, \tilde M^{t-1}}}

\newcommand\pit[1]{\ensuremath{\pi_{[#1]}}}

\newcommand\css[2]{#1[#2]=\tilde M_{#2}}
\newcommand\csson[1]{\bM[\by_{on}^{#1}]=\tilde M_{on}^{#1}}

\newcommand\ehistto[1]{\ensuremath{{#1}_{on}^t, \tilde M_{on}^t}}
\newcommand\ehisttto[1]{\ensuremath{{#1}_{on}^{t-1}, \tilde M_{on}^{t-1}}}

\newcommand{\cG}{\mathcal{G}} 
\newcommand{\cE}{\mathcal{E}} 
\newcommand{\cV}{\mathcal{V}} 
\newcommand{\cN}{\mathcal{N}} 
\newcommand{\cM}{\mathcal{M}} 
\newcommand{\cS}{\mathcal{S}} %

\newcommand{\bM}{\mathbf{M}} 
\newcommand{\bB}{\mathbf{B}} 
\newcommand{\bN}{\mathbf{N}} 
\newcommand{\bx}{\mathbf{x}} 
\newcommand{\bc}{\mathbf{c}} 
\newcommand{\br}{\mathbf{r}} 
\newcommand{\by}{\mathcal{Y}} 
\newcommand{\byo}{\mathcal{Y}_{on}} 
\newcommand{\bys}{\mathcal{Y}_{*}} 
\newcommand{\bz}{\mathcal{Z}} 
\newcommand{\ba}{\mathcal{A}} 
\newcommand{\bb}{\mathcal{B}} 
\newcommand{\bw}{\mathbf{w}} 
\newcommand{\bX}{\mathbf{X}} 

\newcommand{\obsE}[1]{\ensuremath{\mathcal{E}_{#1}^o}}
\newcommand{\tMo}{\tilde{M}_{on}} 

\newcommand{\hth}{\hat{\theta}} 
\newcommand{\xet}{\bx_e^{\top}} 
\newcommand{\xe}{\bx_e} 
\newcommand{\ora}{{\tt{Greedy}}}

\makeatletter
\newenvironment{breakablealgorithm}
{
	\begin{center}
		\refstepcounter{algorithm}
		\hrule height.8pt depth0pt \kern2pt
		\renewcommand{\caption}[2][\relax]{
			{\raggedright\textbf{\ALG@name~\thealgorithm} ##2\par}%
			\ifx\relax##1\relax 
			\addcontentsline{loa}{algorithm}{\protect\numberline{\thealgorithm}##2}%
			\else 
			\addcontentsline{loa}{algorithm}{\protect\numberline{\thealgorithm}##1}%
			\fi
			\kern2pt\hrule\kern2pt
		}
	}{
		\kern2pt\hrule\relax
	\end{center}
}
\makeatother

\usepackage{lmodern}
\usepackage{listings}
\usepackage{amssymb}
\usepackage{xcolor}
\usepackage{textcomp}
\usepackage{lipsum}

\OneAndAHalfSpacedXI

\usepackage{natbib}
\bibpunct[, ]{(}{)}{,}{a}{}{,}%
\def\bibfont{\small}%
\TheoremsNumberedThrough     
\ECRepeatTheorems

\EquationsNumberedThrough    

\begin{document}
	
	
	

	\TITLE{Beyond Adaptive Submodularity: Adaptive Influence Maximization with Intermediary Constraints}
	
	\ARTICLEAUTHORS{%
		\AUTHOR{Shatian Wang, Zhen Xu, Van-Anh Truong}
		\AFF{Department of Industrial Engineering and Operations Research, Columbia University, New York, NY 10027, \EMAIL{sw3219@columbia.edu} 
			\EMAIL{zx2235@columbia.edu}, \EMAIL{vt2196@columbia.edu}}
	} 
	
	\ABSTRACT{We consider a brand with a given budget that wants to promote a product over multiple rounds of influencer marketing. In each round, it commissions an influencer to promote the product over a social network, and then observes the subsequent diffusion of the product before adaptively choosing the next influencer to commission. This process terminates when the budget is exhausted. We assume that the diffusion process follows the popular Independent Cascade model. We also consider an online learning setting, where the brand initially does not know the diffusion parameters associated with the model, and has to gradually learn the parameters over time.
	
Unlike in existing models, the rounds in our model are correlated through an intermediary constraint: each user can be commissioned for an unlimited number of times. However, each user will spread influence without commission at most once. Due to this added constraint, the order in which the influencers are chosen can change the influence spread, making obsolete existing analysis techniques that based on the notion of adaptive submodularity.  We devise a sample path analysis to prove that a greedy policy that knows the diffusion parameters achieves at least $1-1/e - \epsilon$ times the expected reward of the optimal policy.

In the online-learning setting, we are the first to consider a truly adaptive decision making framework, rather than assuming independent epochs, and adaptivity only within epochs.  Under mild assumptions, we derive a regret bound for our algorithm.  In our numerical experiments, we simulate information diffusions on four Twitter sub-networks, and compare our UCB-based learning algorithms with several baseline adaptive seeding strategies. Our learning algorithm consistently outperforms the baselines and achieves rewards close to the greedy policy that knows the true diffusion parameters. 
	}%
	
	
%
	\maketitle
	
	%
	
	
	\section{Introduction}
	Influencer marketing on social networks, such as Facebook, Twitter, and Instagram, has become the fastest growing and most cost-effective form of marketing across many categories.  Brands partner with influential social network users including bloggers, YouTubers, and Instagrammers to get their endorsement.  These influencers post original marketing content for the brand on their social network accounts that have massive followings.  Their followers view the sponsored posts, get informed of the product, and can further share the posts to their social network. Recursively, the sharing activities enable a \textit{diffusion process} in the social network: starting from the influencers, information about the promoted product will spread to a larger audience.
	
	In this paper, we consider a scenario where a brand, with a given budget $B$, wants to promote a product over multiple rounds of influencer marketing. Each user $v$ is associated with a reward $\br(v) \in [r,1]$, where $r$ is a positive constant. The brand has a goal of maximizing the sum of rewards over \textit{distinct} users in the social network who have been \textit{activated} (i.e., influenced) during the process. In each round, it \textit{seeds} \textit{one} influencer $u$ in the network by paying a commission $\bc(u)$. It then observes the subsequent diffusion process from $u$, and in the next round, \textit{adaptively} seeds the next influencer. This process terminates when the budget $B$ is exhausted. We assume that the diffusion process follows the popular \textit{independent cascade (IC)} model by \cite{Kempe:2003}. We also consider an \textit{online learning} setting: the brand initially does not know the diffusion parameters associated with the model, and has to gradually \textit{learn} the parameters over time.
	
	Our problem is closely related to the adaptive influence maximization problem. However, the existing works all treat the diffusion parameters as given. Also, they  assume that either the rounds are completely independent from each other \citep{MRIM}, or that selecting an influencer for a second time results in zero marginal increase in reward to the advertisers \citep{AdaSubm}. We address a problem between the two extremes, where the rounds are correlated through an \textit{intermediary constraint}: each user can be seeded for an unlimited number of times. However, each can serve as an \emph{intermediary}, i.e., a non-seed user that is spreading influence, at most once.  This constraint expresses the fact that a user who is not given incentive to spread influence will stop sharing the same content after she has voluntarily shared it once. Due to this added constraint, the \emph{order} in which the influencers are seeded can change the end result, making the existing analysis techniques that are based on the notion of \emph{adaptive submodularity}, obsolete.
	
We devise a sample path analysis to prove that a greedy policy that knows the diffusion parameters achieves at least $1-1/e - \epsilon$ times the expected reward of the optimal policy. We then consider an online-learning setting of the problem.  Here, an agent needs to \emph{learn} the diffusion parameters during an adaptive seeding process.   All previous works incorporating learning and decision making are epoch-based  \citep{AdaL1, AdaL2}.  Specifically, the time horizon is divided into $n$ \textit{independent epochs}, in each of which the agent makes $T$ rounds of adaptive decisions. Thus, although the problem is adaptive within an epoch, it is not adaptive over the different epochs.  We propose the first UCB-based learning algorithm for this problem under a fully adaptive setting. That is, an agent conducts $T$ rounds of adaptive seeding while learning the parameters. Our algorithm uses the greedy policy to simultaneously learn and exploit the diffusion parameters in these $T$ rounds.

Under mild assumptions, we derive a regret bound for our algorithm.  More precisely, letting $f_{on}^T$ and $f_{opt}^T$ denoting the $T$-round reward of our online policy and that of the optimal policy, respectively, we prove that $\E(f_{on}^T) \geq (1 - e^{-\beta/\alpha}) \E(f_{opt}^T) - \mathcal{O}(c_\cG \cdot g(|\cE|, |\cV|, T) \cdot \sqrt{T})$. Here, $\alpha$ and $\beta$ are two constants from an approximate greedy policy that will be introduced later, $c_{\cG}$ is a network-dependent constant and $g(|\cE|, |\cV|, T)$ is a function of the number of edges $|\cE|$, number of nodes $|\cV|$, and the total number of rounds $T$. The function $g(|\cE|, |\cV|, T)$ grows logarithmically in $|\cE|, |\cV|$, and $T$. Note that due to the lower bound $r >0$ on the reward of each node, $\E(f_{opt}^T)$ grows at least linearly in $T$. Thus a dependency of the loss of reward, called \emph{regret}, that is due to the error of learning the parameters, on $\sqrt{T}\log(T)$ is non-trivial. In our numerical experiments, we simulate information diffusions on four Twitter sub-networks, and compare our UCB-based learning algorithms with several baseline adaptive seeding strategies, such as maximum degree and CUCB-based approximate greedy \citep{CUCB}. Our learning algorithm consistently outpeforms the baselines and achieves rewards close to the greedy policy that knows the true diffusion parameters. 
	
	\section{Problem Formulation}
	
	We model the structure of a social network using a directed graph $\mathcal{D = (V, E)}$. 
	Each node $v \in \mathcal{V}$ represents a user.  An arc $(u,v) \in \mathcal{E}$ exists if that user $v$ is a \textit{follower} of user $u$. Information can spread from $u$ to $v$. We use ``arc'' and ``edge'' interchangeably to refer to a directed edge in $\mathcal{E}$. We assume that each node $v \in \mathcal{V}$ has an activation cost that is given by $\mathbf{c}(v)$. The cost corresponds to the minimum monetary incentive that is required to induce the user to share new content of any kind with her followers.

	We use the \textit{Independent Cascade Model (IC)} \citep{Kempe:2003} to model the diffusion of influence. In this model, each arc $e = (u,v)$ has an associated \textit{influence probability}, denoted as $w(e)$, that measures the likelihood with which user $u$ successfully influences user $v$. 	Throughout the text, we refer to the function $w$ as \textit{influence probabilities}. IC specifies an influence diffusion process that unfolds in discrete time steps. Initially, all nodes are \textit{unactivated}. In step $0$, a seed set $S$ of users is selected and \textit{activated}. In each subsequent step $s$, each user activated in step $s-1$ has a single chance to activate her followers, or \textit{downstream neighbors}, with success rates equal to the corresponding edge influence probabilities. This process terminates when no more users can be activated. 
	
	We can equivalently think of the IC model as flipping a coin on each edge and observing connected components in the graph with edges corresponding to positive flips \citep{Kempe:2003}. More specifically, after the users in the seed set $S$ are activated,  the environment decides on the binary weight function $\mathbf{w}$ by independently sampling $\mathbf{w}(e) \sim \text{Bern}(w(e))$ for each $e\in \mathcal{E}$. 
	A node $v_2 \in \mathcal{V}\backslash S$ is \textit{activated} if there exists a node $v_1 \in S$ and a directed path $e_1, e_2, ..., e_l$ from $v_1$ to $v_2$ such that $\mathbf{w}(e_i) = 1$ for all $i = 1, ..., l$. 
	
	We assume that we can observe the \textit{edge semi-bandit feedback} of the IC model \citep{CUCB, NIPS2017_6895}. That is, for each edge $e = (u,v) \in \mathcal{E}$, we \textit{observe} edge $e$'s realization, $\mathbf{w}(e)$, if and only if the head node $u$ of the edge was activated during the IC process.

	In the setting of \textit{adaptive influence maximization}, an agent promotes the same content over multiple rounds in a social network with influence probability vector $\bar{w}$. In each round $t$, the agent selects \textit{one} seed user $v_t$ to {activate} by paying $\mathbf{c}(v_t)$.  The content then spreads from the seed user according to the IC model.  The IC process in each round $t$ terminates before the next round $t+1$ begins. 
	
	After each round $t$, the agent observes edge semi-bandit feedback of the IC process in that round. That is, an edge $e$'s realization $\mathbf{w}(e) \in \{0,1\}$ is \emph{observed} in round $t$ if and only if the head node of $e$ is activated during the IC process in round $t$. 

	The agent adapts her selection of seed in each round based on observed feedbacks from previous rounds.  Each distinct activated user $v$ contributes a reward $\br(u)$, and the agent's objective is to maximize the expected sum of rewards of these distinct users who are activated over all rounds, under a global budget constraint.
	
	We can interpret a node activation as a user being exposed to the content and deciding whether to share it or not. For an edge $e = (u,v)$, the activation probability can formally be defined as $\bar{w}(e) = Pr\{$user $u$ decides to share the content and $v$ is informed by $u |$ $u$ is exposed to the content$\}$.
	
	\subsection{Limitations of classical adaptive influence maximization}

	Various authors have investigated simple models of adaptive influence maximization.	 Below, we review these models in the context of our application.  We will show that a generalization of these simple models is needed to realistically capture our application.

	\cite{AdaSubm} were the first to analyze adaptive influence maximization. In their model, once an edge $e$'s realization is first observed in a round $t$, it remains fixed for all future rounds. With this assumption, seeding a node for a second time generates no additional influence. For this reason, we call their model the \emph{single-seeding model}.

	The single-seeding model fails to capture the common practice of paying influencers to promote the same content over multiple rounds. The marketing literature has supplied extensive evidence that \textit{repeated advertising} allows a product to be better remembered and more strongly associated with benefits, thus increasing the likelihood that it is considered by customers \citep{10.2307/3151019, doi:10.1086/209145, doi:10.1086/209225, NOEL2006306}. Another reason that repeated promotions are useful is the time sensitivity of major social network platforms. Take Twitter as an example.  With a large number of new tweets appearing on each users' newsfeed but only a limited amount of time that a user scans each tweet at a time, it is very likely that a piece of content might be un-noticed by the user if it is only promoted once \citep{repetition}.  As a result, in reality, brands seek long-term collaborations with influencers to more effectively instill the product or brand recognition in consumers' minds. As an example, the eco-luxury shoe brand Atelier Alienor commissioned the eco-conscious Instagram influencer Natalie Kay to \textit{regularly} post photos about their sustainable shoes. Such a strategy has proven to be highly effective in practice \citep{long-term1, long-term2}.

	To generalize the single-seeding model, \cite{MRIM} consider a richer setting in which, in each round, the edge realizations are generated anew from that round's chosen seed, independently of previous rounds.  With this relaxation, seeding a node $v$ for a second time can have the effect of activating additional neighbors that the node had failed to activate in the first attempt.  We call this model the \emph{multi-seeding model}. 
	
	While it is more realistic than the single-seeding model, the multi-seeding model still falls short of capturing our applications in the following way.  The model assumes that as long as a node $u$ is activated, it will attempt to spread influence, with or without commissions.  In reality, while a user might be willing to promote the same content repeatedly when she is paid commissions, the same cannot be said of unpaid users.  For example, consider a user $u$ who is not seeded (paid commission) in round 1, but who has already attempted to activate her neighbors in round 1 as an intermediary.  Suppose that in all rounds $2, 3, \ldots,T$, $u$ is not seeded but is exposed to the same content.  It does not appear likely that she would repeat her attempts to activate her followers in these additional rounds independently of her prior history.

The simplified dynamics of the single- and multi-seeding models enable both models to fit into the general framework proposed by \cite{AdaSubm} for adaptive sequential decision making with uncertain outcomes and partial observability.  \cite{AdaSubm}'s framework has one important requirement that \emph{the decisions do not alter the underlying dynamics of the system}. In the single-seeding model, for instance, the system dynamics correspond to the edge realizations $\mathbf{w}$. For each edge $e$ in the graph, $\mathbf{w}(e) = 1$ with probability $\bar{w}(e)$, and 0 otherwise, independently of other edges. The decision of seeding a node in any round reveals the realizations of earlier unobserved edges, but does not alter $\mathbf{w}$.  As for the multi-seeding model, the system dynamics correspond to $T$ independent rounds of edge realizations $\mathbf{w}_1, ..., \mathbf{w}_T$. In each round $t$, $\mathbf{w}_t(e) = 1$ with probability $\bar{w}(e)$, independent of other rounds and other edges. Seeding a node in round $t$ reveals the realizations of some edges in the round, but does not alter $\mathbf{w}_s$ of any round $s \in \{1, ..., T\}$.

	In their general framework, \cite{AdaSubm} define the first notions of \emph{adaptive submodularity} and \emph{adaptive monotonicity} that generalize the classical notion of submodularity and monotonicity for set functions. They further show that when an objective function is adaptive submodular and adaptive monotone, the greedy policy, which in each step myopically chooses the action that maximizes the immediate reward, will achieve some performance guarantee when compared to an optimal policy. The reward functions of the single- and multi-seeding models have both been shown to be adaptive monotone and adaptive submodular.  Thus the performance guarantee of the greedy policy follows directly from the general result of \cite{AdaSubm}.

	\subsection{Beyond adaptive submodularity} \label{sec:beyondAda}
	
	Given the limitations of the single- and multi-seeding models in capturing our application, we are motivated to consider the following more general model which we call \textit{Multi-seeding with Intermediary Constraints}. As we will soon show, the important requirement of \cite{AdaSubm}'s general framework, namely, that decisions do not alter the underlying system dynamics, no longer holds for this more general model. In later sections, we will give detailed explanations on how the former requirement greatly simplifies the performance analysis of different seeding policies.  Without it, we must develop new techniques to tackle the performance analysis of the greedy algorithm.
	
	In our model, each node can be \textit{seeded}, or paid commissions, an unlimited number of times. However, each node can serve as an \emph{intermediary}, i.e., a non-seed node that is spreading influence, for at most one round.  This constraint expresses the fact that, realistically, a user who is not given incentive to spread influence will stop sharing the same content after voluntarily sharing it once.   
	
	Mechanistically, we enforce this constraint by permanently removing for all future rounds, all incoming edges to each node $v$, once that node $v$ has been activated once as a non-seed node. (Note that a node can be activated in multiple rounds. Each node $v$ that is activated at least once in the $T$ rounds will contribute $\mathbf{r}(v)$ to the final objective function).  This forced removal has the effect of disabling node $v$ as an intermediary for all future rounds.  Because we remove all incoming edges to node $v$, rather than all outgoing edges, however, node $v$ could still serve as a seed in future rounds.  We call our model \emph{multi-seeding with intermediary constraints}.

	Because of the edge removals, our seeding decisions now do alter the underlying dynamics of the system.  In each round $t$, we will have an updated graph $\mathcal{G}_{t} = (\mathcal{V}, \mathcal{E}_{t})$, with potentially fewer edges compared to the graph $\mathcal{G}_{t-1}$ of round $t-1$, due to edge removals. For each edge $e \in \mathcal{E}_{t}$, its edge realization in round $t$ is still $\mathbf{w}_t(e) = 1$ with probability $\bar{w}(e)$, and 0 otherwise, independent of other rounds and other edges. However, different seeding decisions in round $t-1$ will potentially result in different edge set $\mathcal{E}_t$ of $\mathcal{G}_t$.

\begin{figure}[htbp]
		\centering
		\includegraphics[width=7.5cm]{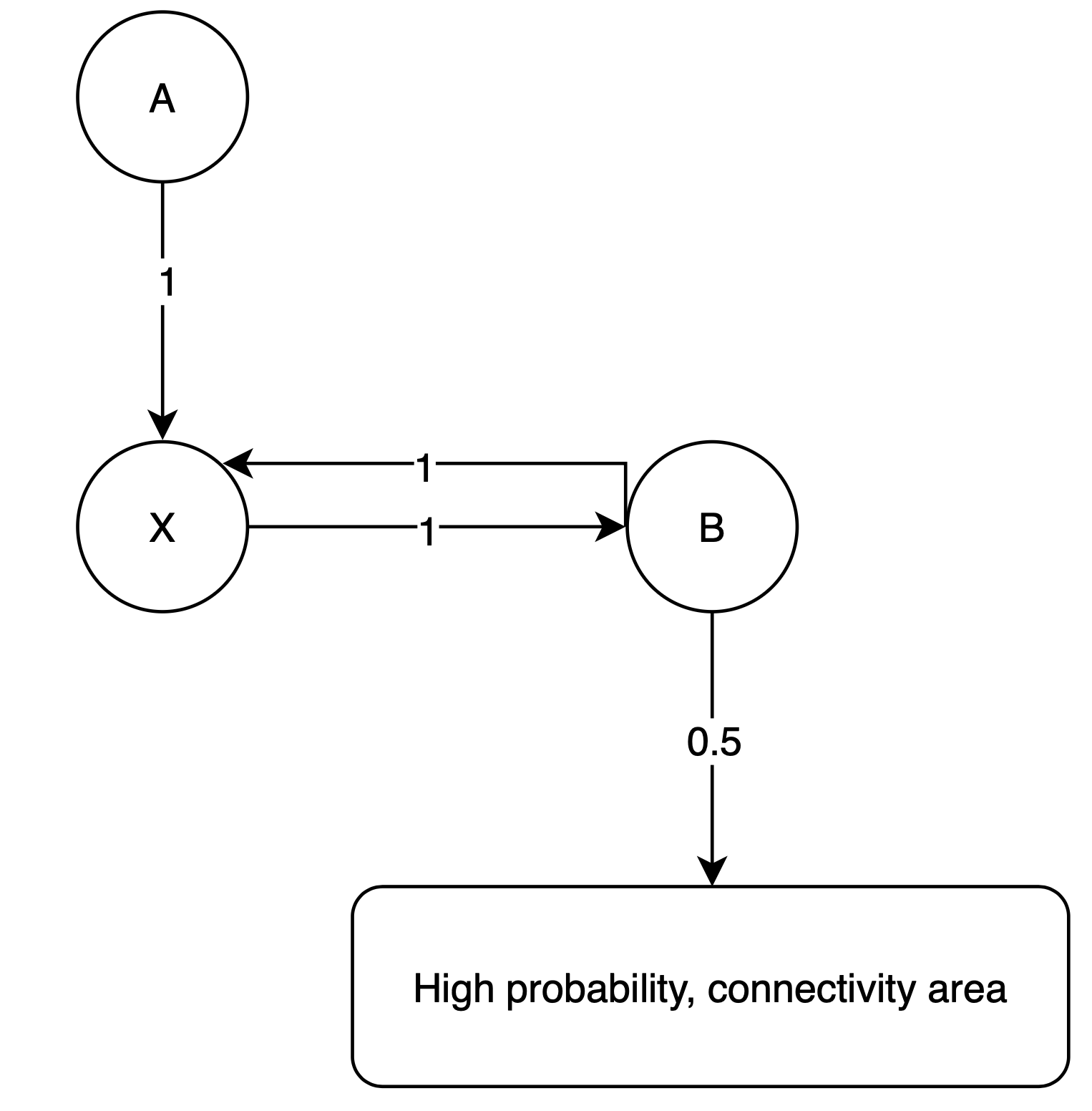}
		\caption{Example showing that seeding order affects the expected reward in the multi-seeding with intermediary constraint model. We want to hit the region with high probability and connectivity. If we seed $A \rightarrow B$, the chance of successfully hitting the region is $0.5 + 0.5 \times 0.5 = 0.75$. On the other hand, if we seed $B \rightarrow A$, the chance of successfully hitting the region is $0.5$. This is because in the round of seeding $B$, node $X$ would serve as an intermediary. Thus in the round of seeding $A$, $X$ would not be reached.}  
		\label{fig:orderMatters0}
	\end{figure}

	Given a fixed seed set $S$, previously for single- and multi-seeding models, the order in which we select the seeds does not affect the final expected reward, since the underlying system dynamics is not changed by the seeding decisions. Now with potentially different edge removals resulting from different seeding decisions, the order of seeding does matter, as illustrated in Figure \ref{fig:orderMatters0}. 
	
	As we will discuss in Section \ref{sct:greedy}, when analyzing the performance guarantee of the greedy policy, we need to analyze a hybrid policy $\pi'$ that first follows a certain number of steps of the greedy policy and then follows the decisions made by an optimal policy. The idea is to use this hybrid policy to link the expected reward of an optimal policy with that of the greedy. A crucial step in making the linkage is to show that the expected reward of $\pi'$ bounds from above the expected reward of an optimal policy. If the order of the decisions does not matter, then it is easy to show that the expected reward of $\pi'$ is the same as the expected reward of a policy $\pi''$ which follows the decision of the optimal policy first, and then follows the greedy policy. Then the problem reduces to proving that the expected reward of $\pi''$ bounds from above the expected reward of an optimal policy. The whole argument is straightforward, since it amounts to showing that the extra seeds will never reduce the expected reward, which is intuitively clear. When order does matter, on the other hand, $\pi'$ and $\pi''$ might no longer have the same expected reward. We thus need to directly compare the hybrid policy with the optimal.

	\subsubsection{Connection to Markov Decision Processes}
	Our problem can be modeled by a Markov Decision Process. Each state of that process can encode, for each node, the number of rounds during which the node has been activated by its upstream neighbors, as well as whether the node has ever been seeded. The action in each round is to choose a node to seed. The action space is thus $\mathcal{V}$. The immediate reward of each round is the sum of node rewards over newly activated nodes in that round. The transition probabilities can be calculated using the edge influence probabilities. However, the size of the state space is exponential in the number of nodes. As a result, finding an optimal policy would take time that is exponential in the number of nodes. We will thus look for polynomial time algorithms with approximation guarantees.
	
	\subsection{Notation and definitions} \label{sct:notations}
	
	To more formally formulate the our adaptive influence maximization with intermediary constraint, we define the following notation. A few definitions to follow are similar to the ones used by \cite{AdaSubm}, but note that the differences in the model will result in different implications for these definitions. For simplicity, we present the analysis assuming uniform node costs. Namely, $\bc(v) = 1 \;\; \forall v\in \cV$. The case with heterogeneous node costs is a simple extension.
	
	\noindent \textbf{Horizon length $T$: } Let $T \in \mathbb{Z}^+$ be a global budget that is given. With the assumption of uniform node costs, we know that the total number of rounds is $T$. 
	
	\noindent \textbf{Seeding sequences:} Let $\cN = \left\{(v_1, ..., v_t) \Big| v_i \in \mathcal{V}, t \leq T\right\}$ be the set of all sequences of nodes with lengths at most $T$. We call each element of size $t$ in $\cN$ a \textit{seeding sequence} of length $t$, i.e., a sequence of $t$ seeds selected one-by-one in order in the first $t$ rounds, and denote it as $\by_t = (v_1, ..., v_t)$. Let $\ba_k = (a_1, a_2, ..., a_k)$, $\bb_l = (b_1, b_2, ..., b_l)$ be two seeding sequences of length $k$ and $l$ respectively. We say that $\ba_k$ is contained in $\bb_l$ (written $\ba_k \subset \bb_l$) if $k \leq l$ and $a_i = b_i \;\; \forall i = 1, ..., l$. We also use $\ba_k@\bb_l$ to denote the concatenated seeding sequence $(a_1, ..., a_k, b_1, ..., b_l)$ of length $k + l$. 
	
	\noindent \textbf{Realization matrix $M$:} A \textit{realization matrix} $M$ is a $|\mathcal{E}| \times T$ binary matrix that completely determines the realizations of the edges in all rounds. Each row of $M$ corresponds to an edge $e \in \mathcal{E}$. We use $M(e, j)$ to denote the $j$th element of row $e$. $M(e,j)$ encodes the realization of edge $e$ (whether it is a success or a failure) the $j$th time that $e$ is \textit{observed} if at all. We use $\cM = \{0,1\}^{|\cE|\times T}$ to denote the set of all realization matrices. Let $P_\cM$ be a probability measure on $\cM$ such that $P_\cM(M) = \prod_{(e, j)}\bar{w}(e)^{M(e,j)}(1-\bar{w}(e))^{1- M(e,j)}$. $\bM$ is a random realization matrix with probability mass given by $P_\cM$.
	
\begin{figure}[htbp]
		\centering
		\includegraphics[width=9.5cm]{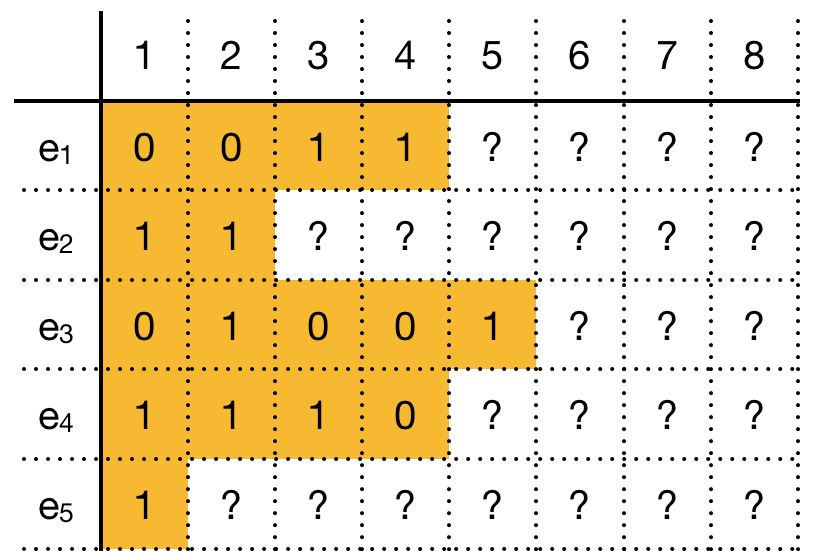}
		\caption{Example of a partial matrix $\tilde M$ for a graph with 5 edges and $T = 8$. The yellow cells correspond to the observed edge realizations. The remaining entries have not yet been observed.}
		\label{fig:M}
	\end{figure}

	\noindent \textbf{Partial matrix $\tilde{M}$:} A \textit{partial matrix} $\tilde{M}$ is a $|\mathcal{E}| \times T$ matrix with entries taking values in $ \{0,1,?\}$. It encodes the ordered realizations \textit{observed} by the agent on each edge, with the question mark denoting unobserved realizations. Each row of $\tilde{M}$ corresponds to an edge $e\in \cE$. We again use $\tilde M(e,j)$ to denote the $j$th element of row $e$. For each row $e$, there exists an index $l,\; 1 \leq l \leq T$, such that $\tilde{M}(e,k) \in \{0,1\} \; \forall k \leq l$, and $\tilde{M}(e, k) = ? \; \forall k > l$. This index $l$ is the number of realizations of edge $e$ the agent has observed. Figure \ref{fig:M} illustrates a partial matrix $\tilde M$.  The yellow cells correspond to the edge realizations the agent has observed. 
	
	Given a realization matrix $M$ and a seeding sequence $\by_t = (v_1, ..., v_t)$, we use $M[\by_t]$ to denote the partial matrix after seeding $\by_t$ under realization matrix $M$. We use $\bM[\by_t]$ to denote the partial matrix associated with $\by_t$ under a random realization matrix $\bM$. We can calculate the probability of observing partial matrix $\tilde M$ after seeding $\by_t$ as $\pr(\bM[\by_t] = \tilde M) = \sum_{M[\by_t] = \tilde M}P_\cM(M)$.
	
	\noindent \textbf{Observed history: } At the beginning of each round $t$, the agent knows the nodes that have been seeded in the previous $t-1$ rounds, i.e., the seeding sequence $\by_{t-1} = (v_1, ..., v_{t-1})$. The agent also has a partial matrix $\tilde{M}_{\by_{t-1}}$ that encodes the observed edge realizations so far. $\tilde M_{\by_{t-1}}$ is a realized sample of $\bM[\by_{t-1}]$. We refer to $\hist{\by_{t-1}}$ as an \textit{observed history} of the first $t-1$ round since it encodes all the information that the agent has gathered during the first $t-1$ rounds. The agent makes the next seeding decision in round $t$ solely based on $\hist{\by_{t-1}}$. 
	
	

	\noindent \textbf{Consistency and sub-history:} Given an observed history $\hist{\by}$ and a realization matrix $M$, if $M[\by] = \tilde M_{\by}$, then we say that the observed history $\hist{\by}$ is \textit{consistent} with the realization matrix $M$. Given two observed histories $\hist{\by}$ and $\hist{\bz}$, if $\by \subset \bz$ and there exists a realization matrix $M$ such that $\tilde M_{\by} = M[\by]$ and $\tilde M_\bz= M[\bz]$, then we say that $\hist{\by}$ is a \textit{sub-history} of $\hist{\bz}$, written $\hist{\by} \subset \hist{\bz}$.
	
	%
	%
	%
	%
	
	\noindent \textbf{Objective function $f(\by, M)$}: $f(\by, M)$ denotes the sum of node rewards over distinct activated users given a seeding sequence $\by$ and a realization matrix $M$. $f(\by, M)$ takes the form $ \sum_{v \in S} \br(v)$ where $S$ is the set of activated nodes. Recall that $\tilde M_\by = M[\by]$ is the partial matrix the agent observes after seeding $\by$ under $M$. A node $v \in \cV$ is activated if and only if $v$ is seeded, i.e., $v\in \by$, or there exists some $e = (u, v) \in \cE$ and $j \in \{1, ..., T\}$ such that $\tilde M_\by(e, j) = 1$. Therefore, $f(\by, M)$ is completely determined by $[\by, \tilde M_\by]$, and thus with a slight abusion of notation, we sometimes write $f(\by, M)$ as $f(\by, \tilde M_{\by})$.
	
	\noindent \textbf{Adaptive policies $\pi$:} An adaptive policy $\pi$ is a mapping from a \textit{subset} of observed histories, $dom(\pi)$, to the node set $\mathcal{V}$. Following a policy $\pi$ means that in each round $t$, we select $v_t = \pi(\hist{\by_{t-1}})$, where $\hist{\by_{t-1}}$ is the observed history of the first $t-1$ rounds. Recall that we use $\ba_k@\bb_l$ to denote the concatenated seeding sequence $(a_1, ..., a_k, b_1, ..., b_l)$ of length $k + l$. The observed history after selecting $v_t$ in round $t$ is $[\by_{t}, \tilde \bM_{\by_t}]$, where $\by_{t} = \by_{t-1}@(v_t)$ and $\tilde \bM_{\by_t}$ is a random partial matrix distributed as $\bM[\by_t]\Big | \left(\css{\bM}{\by_{t-1}}\right)$. Due to our intermediary constraints, edges pointing into the nodes that have helped spread influence as intermediary nodes are removed after each round. The realizations of edges in different rounds for the ones that remain in the network are still independent. As a result, even if $v_t$ has already been seeded in previous rounds, seeding it for another time can still result in potential increase in reward.  The policy terminates in round $\tau$ when $\hist{\by_{\tau-1}} \notin dom(\pi)$. 
	
	We use $\pi(M)$ to denote the complete seeding sequence decided by $\pi$ under realization matrix $M$. $\pi(M)$ has the form $(v_1, v_2, ..., v_{\tau})$, where $\tau$ is the stopping time of the policy. The reward of an adaptive policy $\pi$ under realization matrix $M$ is given by $f(\pi(M), M)$. We further represent the expected reward of $\pi$ as $f_{avg}(\pi) = \mathbb{E}(f(\pi(\bM), \bM))$, where $\bM$ is a random realization matrix with distribution $P_\cM$. The expectation is with respect to both the randomness of $\bM$ and the potential randomness of $\pi$ if it is a randomized policy.
	
	\noindent \textbf{Policy truncation:} an adaptive polict $\pi$ might run for more than $T$ rounds and we thus introduce a notion of policy truncation. We use $\pit{k}$ to denote a policy that runs for at most $k$ rounds. 
	\begin{definition} Given a policy $\pi$, its \emph{level-$k$ truncation $\pit{k}$} has domain $$\left\{\hist{\by} \in dom(\pi): |\by| < k \right\}$$ and agrees with $\pi$ everywhere in its domain. As a result, $\pit{k}$ terminates in at most $k$ rounds.
	\end{definition}

	\noindent \textbf{Optimal policy $\pi^*$:} With the notation above, the objective of our problem is to find an optimal policy $\pi^* \in \text{argmax}_{\pi} f_{avg}(\pi_{[T]})$ for a given budget $T \in \mathbb{Z}^+$.

	\section{Offline Problem and Performance of the Greedy Policy} \label{sct:greedy}
	
	Before exploring the learning paradigm of our multi-round adaptive influence maximization problem with intermediary constraint, we first analyze the offline problem.  More specifically, we assume that the edge influence probabilities are given. We prove a lower bound on the expected reward achieved by the greedy policy relative to the optimal policy. Later we will analyze an online learning algorithm for the problem with unknown influence probabilities using the insights developed in this section.
	
	To obtain the lower bound of the greedy policy's performance in terms of the optimal policy's performance, we analyze a hybrid policy that seeds the first $t$ nodes selected by the greedy policy and then selects the $T$ seeds following the optimal policy. As briefly mentioned in Section \ref{sec:beyondAda} and will be discussed in greater details in Section \ref{sec:link}, we use this hybrid policy as a bridge: on the one hand, we upper bound the expected reward of the optimal policy by the reward of this hybrid policy; on the other hand, we upper bound the reward of the hybrid policy by greedy policy's marginal benefital rewards. Since the seeding order can change the final result under intermediary constraints, the first bound cannot be directly established as in the case of the single- and multi-seeding models. We thus devise a novel sample-path analysis to show that the first bound holds under every sample path, i.e., every realization matrix $M$.

	\subsection{Expected marginal gain of seeding given observed history}
	
	\begin{definition} The \textit{expected marginal gain} of seeding $v$ given an observed history $\hist{\by}$ is  \begin{equation}\begin{split}\Delta(v|\ehist{\by}) &= \E\left[f(\by @(v), \bM) - f(\by, \bM) \Big| \css{\bM}{\by}\right]\\ &= \E\left[f(\by @(v), \bM)\Big | \css{\bM}{\by}\right] - f(\by, \tilde M_{\by}).\end{split}\end{equation}
	\end{definition}

	Below, we prove two important properties of $\Delta(v|\ehist{\by})$. They are similar to the notions of adaptive monotonicity and adaptive submodularity proposed by \cite{AdaSubm}.  The main difference is that the first argument of our objective function $f(\cdot, \cdot)$ is an ordered sequence that is meant to capture the importance of the seeding order, while in theirs, it is a set.
	
	\begin{lemma}
		For all $v \in \mathcal{V}$ and observed history $\hist{\by}$ such that $\pr(\css{\bM}{\by}) = \sum_{M: \css{M}{\by}} P_\cM(M) >0$, we have $\Delta(v|\ehist{\by}) \geq 0$.
	\end{lemma}
	
	\proof{Proof.}
	
	For any realization matrix $M$ such that $\css{M}{\by}$, each node $u$ that has been activated during the observed history $\hist{\by}$ and thus contributing $\br(u)$ to $f(\by, \tilde M_{\by})$ will also contribute $\br(u)$ to $f(\by@(v), M)$. Since $\br \geq 0$, $\E\left[f(\by@(v), \bM)\Big| \css{\bM}{\by}\right] - f(\by, \tilde M_{\by}) \geq 0$.\Halmos
	
	\endproof
	
	\begin{lemma} \label{lemma:2}
		For all $v \in \mathcal{V}$ and $\hist{\by}, \hist{\bz}$ such that $P(\css{\bM}{\by}) >0, P(\css{\bM}{\bz}) >0$, and $\hist{\by} \subset \hist{\bz}$, we have $\Delta(v|\ehist{\by}) \geq \Delta(v|\ehist{\bz})$. 
	\end{lemma}
	
	\proof{Proof.} 
	Let $\mathcal{V}_1$ be the set of nodes that have been activated under $\hist{\by}$, and $\mathcal{G}_\by = (\mathcal{V}, \mathcal{E}_\by)$ be the updated graph after seeding $\by$ and observing $\hist{\by}$ (that is, after removing all the incoming edges to nodes that have served as intermediary in spreading influence for $K$ rounds under $\hist{\by}$). Let $\mathcal{V}_2$ be the set of nodes that are not in $\mathcal{V}_1$ but have been activated under $\hist{\bz}$, and $\mathcal{G}_\bz$ be the graph after observing $\hist{\bz}$. Let $\mathcal{V}_3 = \mathcal{V} \backslash (\mathcal{V}_1 \cup \mathcal{V}_2)$.
	
	We have \begin{align}\Delta(v|\ehist{\by}) &= \E[f(\by@(v), \bM)\Big| \css{\bM}{\by}] - f(\by, \tilde M_{\by})\\
		& = \sum_{v' \in \mathcal{V}_2 \cup \mathcal{V}_3} \br(v')\cdot \pr(v' \text{ is activated} | v, \mathcal{G}_\by)\\
		& = \sum_{v' \in \mathcal{V}_2} \br(v')\cdot \pr(v' \text{ is activated} | v, \mathcal{G}_\by) + \sum_{v' \in \mathcal{V}_3} \br(v')\cdot \pr(v' \text{ is activated} | v, \mathcal{G}_\by) \label{eq:split}.\end{align}
	
	Similarly, \begin{align}\Delta(v|\ehist{\bz}) &= \E\left[f(\bz@(v), \bM)\Big| \css{\bM}{\bz}\right] - f(\bz, \tilde M_\bz)\\
		& = \sum_{v' \in \mathcal{V}_3} \br(v')\cdot \pr(v' \text{ is activated} | v, \mathcal{G}_\bz).\end{align}
	
	Since $\mathcal{G}_\bz$ is a subgraph of $\mathcal{G}_\by$ (with possibly fewer edges), for each $v' \in \mathcal{V}$, we have that $$\pr(v' \text{ is activated} | v, \mathcal{G}_\bz) \leq \pr(v' \text{ is activated} | v, \mathcal{G}_\by).$$ As a result, \begin{align} \Delta(v|\ehist{\bz}) &= \sum_{v' \in \mathcal{V}_3} \br(v')\cdot \pr(v' \text{ is activated} | v, \mathcal{G}_\bz)\\ & \leq \sum_{v' \in \mathcal{V}_3} \br(v')\cdot \pr(v' \text{ is activated} | v, \mathcal{G}_\by)\\ &\leq \sum_{v' \in \mathcal{V}_2} \br(v')\cdot \pr(v' \text{ is activated} | v, \mathcal{G}_\by) + \sum_{v' \in \mathcal{V}_3} \br(v')\cdot \pr(v' \text{ is activated} | v, \mathcal{G}_\by)\\ &=\Delta(v|\ehist{\by}).\Halmos\end{align} 
	\endproof
	
	\subsection{Greedy policy and $\alpha$-approximate greedy policy} 
	
	We define a \textit{greedy policy} $\pi$ to be the one that in each round $t$ selects a node $v_t$ to maximize $\Delta(v|\ehist{\by_{t-1}})$, where $\hist{\by_{t-1}}$ is the observed history of the first $t-1$ rounds.  The policy terminates in round $\tau$ when $\Delta(v|\ehist{\by_{\tau-1}}) = 0 \; \forall v \in \cV$. 
	
	Since $\Delta(v|\ehist{\by})$ is computationally intractable to evaluate exactly, we assume that we have access to an \textit{$\alpha, \beta$-approximate greedy policy}:
	
	\begin{definition}
		A policy $\pi$ is an \textit{$\alpha, \beta$-approximate greedy policy} if for all observed history $\hist{\by}$ such that there exists $v \in \mathcal{V}$ with $\Delta(v|\ehist{\by}) > 0$, 		
		$$\pi(\hist{\by}) \in \left\{v: \Delta(v|\ehist{\by}) \geq \frac{1}{\alpha} \max_{v'} \Delta(v'|\ehist{\by})\right\}$$ with probability at least $\beta$.
		Moreover, $\pi$ terminates upon observing any $\hist{\by}$ such that $\Delta(v|\ehist{\by}) = 0$ for all $v \in \mathcal{V}$. When $\alpha = \beta = 1$, $\pi$ is a greedy policy.  
	\end{definition}
	\subsection{Performance guarantee of $\alpha$-approximate greedy algorithm} \label{sec:link}
	We want to prove the following hypothesis:
	
	\begin{hypothesis}\label{hyp:greedy_guarantee} Let $\pi$ be an $\alpha, \beta$-approximate greedy policy. For an optimal policy $\pi^*$ and positive integers $l$ and $k$, $f_{avg}(\pi_{[l]}) \geq (1-e^{-\beta l/\alpha k})f_{avg}(\pi^*_{[k]})$. 
	\end{hypothesis}
	
	If we set $l = k = T$, the hypothesis tells us that an $\alpha, \beta$-approximate greedy policy achieves in expectation at least $1-e^{-\beta/\alpha}$ the influence spread achieved by an optimal policy, with respect to level-$T$ truncations. 
	
	To prove Hypothesis \ref{hyp:greedy_guarantee}, we need to link the expected reward of the seeding sequence $\pi_{[l]}(\bM)$ with that of the seeding sequence $\pi^*_{[k]}(\bM)$, both under a random realization matrix $\bM$. To do so, we analyze the expected reward of a \textit{hybrid} seeding sequence $\pi_{[h]}(\bM)@\pi^*_{[k]}(\bM)$ for each $h \in \{1, ..., l-1\}$. We will first show that the expected reward of seeding $\pi_{[h]}(\bM)@\pi^*_{[k]}(\bM)$ is an upper bound on the expected reward of seeding only $\pit{k}^*(\bM)$.  We call this result Linkage 1. On the other hand, we can show that the expected reward of seeding the hybrid sequence $\pit{h}(\bM)@\pit{k}^*(\bM)$ is bounded above by the expected reward of seeding $\pit{h}(\bM)$ plus some additional terms that can be related to the greedy seed selection scheme.  We call this second result Linkage 2.  
	
	We will first establish Linkage 1.  That is, for all $h, \; 0 \leq h < l$, \begin{equation}f_{avg}(\pi^*_{[k]}) = \E[f(\pit{k}^*(\bM), \bM)] \leq \E[f(\pit{h}(\bM)@\pit{k}^*(\bM),  \bM)].\label{eq:first_link}\end{equation}
	
	Note that if the seeding order does not matter, then 
	\begin{equation}\E[f(\pit{h}(\bM)@\pit{k}^*(\bM),  \bM)] = \E[f(\pit{k}^*(\bM)@\pit{h}(\bM), \bM)].\label{eq:switch_order}\end{equation}
	
	\cite{AdaSubm} show that when an objective  function satisfies the general adaptive monotonicity according to their definition, then 
	\begin{equation}\E[f(\pi_1(\bM), \bM)] \leq \E[f(\pi_1(\bM)@\pi_2(\bM), \bM)]\label{eq:adaMonEquiv}\end{equation} 
	for any two policies $\pi_1$ and $\pi_2$. The statement above is the implication of their result expressed in our notations and setup. \cite{AdaSubm} prove their result by contradiction.  The proof is relatively straightforward.  Note that in our setting, \eqref{eq:adaMonEquiv} is very intuitive to see, since seeding extra nodes never reduces the final reward.  However, \eqref{eq:first_link} is less than obvious.
	
	\cite{AdaSubm} then combine \eqref{eq:adaMonEquiv} together with \eqref{eq:switch_order} to show \eqref{eq:first_link}. Thus, inequality \eqref{eq:first_link} is proved in single- and multi-seeding models.  However, in our setup, the seeding order turns out to be crucial. We  need to prove that inequality \eqref{eq:first_link} holds without relying on \eqref{eq:switch_order}. We do so using the following sample path argument: for each realization matrix $M$ and policies $\pi_1$ and $\pi_2$,  $f(\pi_1(M), M) \leq f(\pi_2(M)@\pi_1(M), M)$.

	\begin{theorem}\label{thm:greedy_guarantee_step1}
		Fix a $|\mathcal{E}|\times 2T$ binary realization matrix $M$. Let $\pi_1, \pi_2$ be any two policies that both terminate within $T$ rounds. Then $f(\pi_1(M), M) \leq f(\pi_2(M)@\pi_1(M), M)$.
	\end{theorem}
	\proof{Proof.}
	Let $\by_{t_1} = \pi_1(M)$ $= (v_1, ..., v_{t_1})$, $\bz_{t_2} = \pi_2(M)$ $= (u_1, ..., u_{t_2})$, $\bz_{t_2}@\by_{t_1} = \pi_2(M)@\pi_1(M)$ $= (u_1, ..., u_{t_2}, v_1, ..., v_{t_1})$. 
	With $\tilde M(e) = \tilde M(u, v)$ denoting the row corresponding to edge $e = (u,v)$ in partial matrix $\tilde M$, we first show that $M[\by_{t_1}](e) \subset M[\bz_{t_2}@\by_{t_1}](e) \;\; \forall e \in \mathcal{E}$. Or if we use $|\tilde M(e)|$ to denote the number of non-question mark entries of $\tilde M(e)$, then equivalently we want to show that $|M[\by_{t_1}](e)| \leq |M[\bz_{t_2}@\by_{t_1}](e)| \;\;\forall e \in \mathcal{E}$. We do so by induction on $t_1$. For simplicity, we use $\by_{s}$ to denote $(v_1, ..., v_s)$ with $s = 1, ..., t_1$. Similarly, we use $\bz_{s}$ to denote $(u_1, ..., u_s)$ with $s = 1, ..., t_2$.
	
	In the base case $t_1 =0$, the containment obviously holds for any $t_2$. Now assume that the containment holds for $t_1 \leq s_1$ and any $t_2$. We now show that the containment also holds for $t_1 = s_1+1$ and any $t_2$. Suppose to the contrary that $M[\by_{s_1+1}]$ is not contained in $M[\bz_{t_2}@\by_{s_1+1}]$.  Then there exists an edge $e = (A, B)$ such that $|M[\by_{s_1+1}](e)| > |M[\bz_{t_2}@\by_{s_1+1}](e)|$. 
	
	If $|M[\by_{s_1+1}](e)| = |M[\by_{s_1}](e)|$, then by the inductive hypothesis, $|M[\by_{s_1+1}](e)| = |M[\by_{s_1}](e)| \leq |M[\bz_{t_2}@\by_{s_1}](e)| \leq |M[\bz_{t_2}@\by_{s_1+1}](e)|$, which is a contradiction. Therefore $e$ is observed in the last round of seeding $\by_{s_1+1}$, and is not observed in the last round of seeding $\bz_{t_2}@\by_{s_1+1}$. Furthermore, $|M[\by_{s_1}](e)| = |M[\bz_{t_2}@\by_{s_1}](e)|$. 
	
	If $v_{s_1+1} = A$ where $A$ is the head node in edge $e =(A,B)$ specified above, then $e$ would have been observed in the last round of seeding $\bz_{t_2}@\by_{s_1+1}$, a contradiction. Thus, $v_{s_1+1} \neq A$. Also, all observations of $e$ during the process of seeding $\by_{s_1}$ must have been due to seeding $A$, since otherwise, $A$ would have served as an intermediary node and thus when seeding $v_{s_1+1}$, there will not be any path connecting to $A$. 
	
	Since $e$ is observed in the last round of seeding $\by_{s_1+1}$, there exists a path $P = \{w_1, ..., w_l\}$ such that  $w_1 = v_{s_1+1}$ and $w_l = A$, and $(w_i, w_{i+1}) = 1$ for all $i = 1, ..., l-1$ when seeding $v_{s_1+1}$ following $\by_{s_1}$. We also know that for each of these edges, all previous observations (before the 1) are all 0's. This is because if any of them had been 1, its tail node would have served as an intermediary, and $P$ would have no longer existed when $v_{s_1+1}$ was seeded following $\by_{s_1}$. Furthermore, all the previous observations of these edges (except for $(w_1, w_2)$) must have been due to directly seeding their head nodes, since otherwise, the head node would have served as an intermediary, and the path would have been disconnected in the last round of seeding $\by_{s_1+1}$.
	
	\begin{figure}[htbp]
		\centering
		\includegraphics[width=9cm]{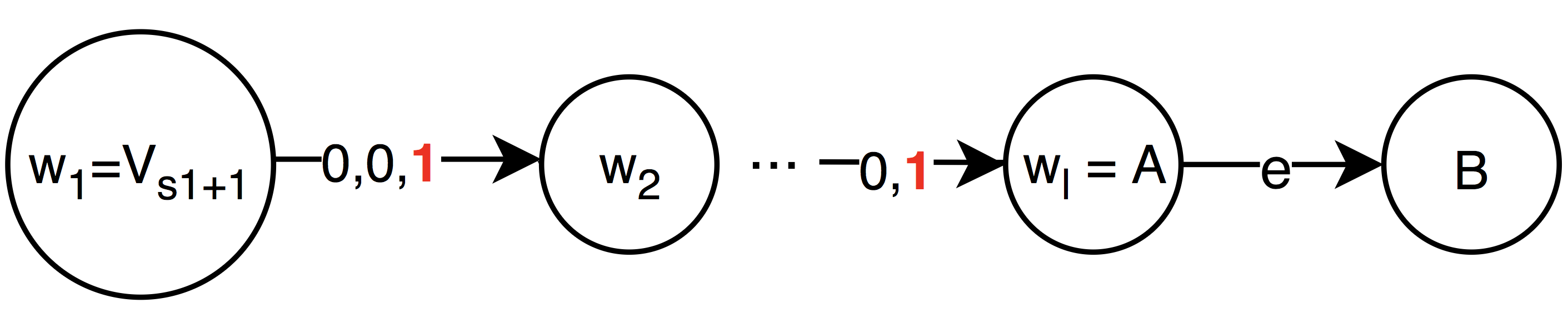}
		\caption{In the last round of seeding $v_1, ..., v_{s_1+1}$, $e$ is activated through the path.}
		\label{fig:illus1}
	\end{figure}
	
	Since $e$ is not observed in the last round of seeding $\bz_{t_2}@\by_{s_1+1}$, this path cannot be all 1's in the last round. This might happen if some edge $e' = (w_i, w_{i+1})$ in $P$ has not reached its first 1 in the matrix $M$ and would be observed to be 0 in the round of seeding $v_{s_1+1}$ following $\bz_{t_2}@\by_{s_1}$. But by the inductive hypothesis, $M[\bz_{t_2}@\by_{s_1}](e') \supset M[\by_{s_1}](e')$, and the latter includes the last 0 observation before the first 1 of edge $e'$. This is a contradiction. Therefore, it can only happen if $M[\bz_{t_2}@\by_{s_1}](e')$ contains the first 1 for some $e' \in P$. If $e' = (w_{l-1}, w_l)$, then when $e'$ is observed to be 1, $e$ will be observed too. But we have already argued that all observations of $e$ during the process of seeding $\by_{s_1}$ must have been due to seeding $A$, and $|M[\by_{s_1}](e)| = |M[\bz_{t_2}@\by_{s_1}](e)|$. This equality will not hold if in seeding $\bz_{t_2}@\by_{s_1}$, $e$ is also observed not through directly seeding $A$. We have thus established that $|M[\by_{s_1}](w_{l-1}, w_l)| = |M[\bz_{t_2}@\by_{s_1}](w_{l-1}, w_l)|$. Therefore, following the same argument, we have that $|M[\by_{s_1}](w_{l-2}, w_{l-1})| = |M[\bz_{t_2}@\by_{s_1}](w_{l-2}, w_{l-1})|$, ..., $|M[\by_{s_1}](w_{1}, w_2)| = |M[\bz_{t_2}@\by_{s_1}](w_{1}, w_{2})|$. As a result, $P$ will be all 1's in the last round of seeding $\bz_{t_2}@\by_{s_1+1}$. This is a contradiction. 
	
	Now that we have established that $M[\by_{t_1}](e) \subset M[\bz_{t_2}@\by_{t_1}](e) \;\; \forall e \in \mathcal{E}$, we proceed to show $f(\pi_1(M), M) \leq f(\pi_2(M)@\pi_1(M), M)$. This fact is very easy to see. Given any seeding sequence $\by$ and realization matrix $M$, each node $v \in \mathcal{V}$ is activated and thus contributing $\br(v)$ towards $f(\by, M)$ if and only if either $v \in \by$ or $\sum_{(w,v) \in \mathcal{E}} ||M[\by](w,v)||_1 > 0$. Since $\by_{t_1} \subseteq \bz_{t_2}@\by_{t_1}$ and $M[\by_{t_1}](e) \subset M[\bz_{t_2}@\by_{t_1}](e) \;\; \forall e \in \mathcal{E}$, we can conclude that $f(\pi_1(M), M) \leq f(\pi_2(M)@\pi_1(M), M)$. \Halmos 
	
	To establish linkage 2, we prove the following theorem. The structure of the proof is similar to that of Theorem 38 in \citep{AdaSubm}. However, the original proof assumes that seeding decisions do not alter the underlying system dynamics and thus the order of the decisions does not matter, as long as the set of decisions is fixed. We extend the proof to work for our more general scenario, where previous seeding decisions might alter the dynamics of later rounds.
	
	Theorem \ref{thm:greedy_guarantee_step2} below upper bounds the expected increase in reward in seeding $\pi(\bM)$, i.e., the seeding sequence determined by any policy $\pi$ under a random realization matrix $\bM$, after seeding $\by$ and observing $\bM[\by] = \tilde M_{\by}$. We refer to this expected increase in reward as \textit{quantity 1}. The upper bound is the expected cost of seeding $\pi(\bM)$ conditioned on $\bM[\by] = \tilde M_{\by}$ times the expected increment in reward of seeding the node returned by the exact greedy algorithm. We refer to this upper bound as \textit{quantity 2}. As will be detailed later, this result allows us to establish linkage 2, that is, the expected reward of seeding the hybrid sequence $\pit{h}(\bM)@\pit{k}^*(\bM)$ can be bounded above by the expected reward of seeding $\pit{h}(\bM)$ plus some additional terms that can be related to the greedy seed selection scheme.
	
	At a high level, the proof uses an optimization problem to link quantity 1 and quantity 2. The feasible solutions to the optimization problem are probability distributions on the set of seeding sequences of lengths at most $T$. The proof first constructs a probability distribution that is feasible to the optimization problem, and shows the objective value with respect to this feasible solution is an upper bound on quantity 1. Since it is a maximization problem, the optimal objective value of the optimization problem upper bounds quantity 1. Then we proceed to show that the first constraint of the optimization problem allows us to upper bound the optimal objective value by quantity 2.

	\begin{theorem}\label{thm:greedy_guarantee_step2}  
		Suppose we have the observed history $\hist{\by}$ after selecting $\by$. Let $\pi$ be any policy.
		Let $Z$ be the solution to the optimization problem
		\begin{equation*}
			\begin{array}{ll@{}ll}
				Z= \max\limits_x  &\sum_{\bz \in \cN} x(\bz) \sum_{v \in \bz\backslash \by} \Delta(v|\ehist{\by})\\
				\text{s.t.} &  \sum_{\bz \in \cN} x(\bz) \sum_{v \in \bz\backslash \by} \mathbf{c}(v) \leq \E\left[\mathbf{c}(\pi(\bM))\Big|\css{\bM}{\by}\right]\\
				& 0 \leq x \leq 1.
			\end{array}
		\end{equation*}
		Then  
		\begin{equation} 
		\begin{split}
		&\E\left[f(\by@\pi(\bM), \bM) - f(\by, \bM)\Big| \css{\bM}{\by}\right]\label{eq:marginal1}\\ 
		& = \E\left[f(\by@\pi(\bM), \bM)\Big| \css{\bM}{\by}\right] - f(\by, \tilde M_{\by})\\
		& \leq Z\leq \E\left[\mathbf{c}(\pi(\bM))\Big|\css{\bM}{\by}\right] \text{max}_v \left(\frac{\Delta(v|\ehist{\by})}{\mathbf{c}(v)}\right).
		\end{split} 
		\end{equation}		
	\end{theorem}

	\proof{Proof.} Recall that $\cN = \{(v_1, ..., v_t), v_i \in \mathcal{V}, t \leq T\}$ is the set of all sequences of nodes with lengths at most $T$. Fix a observed history $\hist{\by}$. Let $\bM$ be a random realization matrix.
	Consider a policy $\pi'$ that runs under $\bM$: $\pi'$ seeds nodes following the order of $\by = (v_1, ..., v_t)$ one by one; it will terminate in round $k$ for $k \leq t$ if $\bM(v_1, ..., v_{k-1}) \not\subset \tilde M_{\by}$; if $\css{\bM}{\by}$, then $\pi'$ proceeds after seeding $\by$ to seed nodes in $\pi(\bM)$ one by one. 
	Note that the expected marginal benefit of running the $\pi$ portion of $\pi'$, given $\css{\bM}{\by}$ equals \eqref{eq:marginal1}.
	
	Now for any seeding sequence $\bz \in \cN$, let $p(\bz) = Pr(\pi'(\bM) = \bz\Big|\css{\bM}{\by})$ be the probability that $\bz$ is the seeding sequence selected by $\pi'$ given $\css{\bM}{\by}$. By the construction of $\pi'$, we know that conditioned on $\css{\bM}{\by}$, $\bz$ must agree with $\by$ in the first $t$ elements to have positive probability $p(\bz)$. Now let $\bz = (v_1, ..., v_t, u_1, ..., u_s)$ and for $k = 1, ..., s$, let $\bz_k = (v_1, ..., v_t, u_1, ..., u_k)$. Conditioned on $\css{\bM}{\by}$ and $\by@\pi(\bM) = \pi'(\bM) = \bz$, when $u_k$ is seeded in round $t+k$, its expected marginal contribution to $\E\left[f(\by@\pi(\bM), \bM) - f(\by, \bM) \Big| \css{\bM}{\by}, \pi'(\bM) = \bz\right]$ is $\Delta(u_k|\bz_{k-1}, \bM[\bz_{k-1}])$. Since $\hist{\by} \subset [\bz_{k-1}, \bM[\bz_{k-1}]]$, by Lemma \ref{lemma:2}, we have that $\Delta(u_k|\bz_{k-1}, \bM[\bz_{k-1}]) \leq \Delta(u_k|\ehist{\by})$ for $k = 1, ..., s$. Therefore, $$\E\left[f(\by@\pi(\bM), \bM) - f(\by, \bM)\Big| \css{\bM}{\by}, \pi'(\bM) = \bz\right] \leq \sum_{v \in \bz\backslash \by} \Delta(v|\ehist{\by}).$$ As a result,  $$\E\left[f(\by@\pi(\bM), \bM) - f(\by, \bM)\Big| \css{\bM}{\by}\right] \leq \sum_{j \in \cN} p(\bz) \sum_{v \in \bz\backslash \by} \Delta(v|\ehist{\by}).$$ Clearly, by the definition of $p(\bz)$, $p$ also satisfies the two constraints of the LP. As a result, $p$ is a feasible solution to the maximization LP. Thus, $Z$ is bounded below by $\E\left[f(\by@\pi(\bM), \bM) - f(\by, \bM) \Big| \css{\bM}{\by}\right]$. So far, we have proved the first inequality in \eqref{eq:marginal1}. 
	
	The second inequality in \eqref{eq:marginal1} is easy to see.  For any feasible solution $x$ to the LP,  
	
	\begin{align}&\sum_{\bz \in \cN} x(\bz) \sum_{v \in \bz\backslash \by} \Delta(v|\ehist{\by})\\ 
		&= \sum_{\bz \in \cN} x(\bz) \sum_{v \in \bz\backslash \by} \mathbf{c}(v) \frac{\Delta(v|\ehist{\by})}{\mathbf{c}(v)}\\
		&\leq \sum_{\bz \in \cN} x(\bz) \sum_{v \in \bz\backslash \by} \mathbf{c}(v) \text{max}_{v'} \left(\frac{\Delta(v'|\ehist{\by})}{\mathbf{c}(v')}\right)\\
		&= \text{max}_{v'} \left(\frac{\Delta(v'|\ehist{\by})}{\mathbf{c}(v')}\right) \sum_{\bz \in \cN} x(\bz) \sum_{v \in \bz\backslash \by} \mathbf{c}(v)\\
		& \leq \text{max}_{v'} \left(\frac{\Delta(v'|\ehist{\by})}{\mathbf{c}(v')}\right) \E\left[\mathbf{c}(\pi(\bM)) \Big|\css{\bM}{\by}\right].
	\end{align}
	
	The last inequality follows from the first constraint of the LP. \Halmos
	\endproof

	With Theorems \ref{thm:greedy_guarantee_step1}, \ref{thm:greedy_guarantee_step2}, we now prove Hypothesis \ref{hyp:greedy_guarantee}:
	
	\begin{theorem} \label{thm:greedyMain}
		Let $\pi$ be an $\alpha,\beta$-approximate greedy policy. For an optimal policy $\pi^*$ and positive integers $l$ and $k$, $f_{avg}(\pi_{[l]}) \geq (1-e^{-\beta l/\alpha k})f_{avg}(\pi^*_{[k]})$. 
	\end{theorem}
	
	\proof{Proof.}
	Without loss of generality, we can assume that $\pi = \pi_{[l]}$ and $\pi^* = \pi^*_{[k]}$. For all $h, \; 0 \leq h < l$, we have from Theorem \ref{thm:greedy_guarantee_step1} that
	
	\begin{equation}f_{avg}(\pi^*) = \E[f(\pi^*(\bM), \bM)] \leq \E[f(\pit{h}(\bM)@\pi^*(\bM),  \bM)],\label{eq:ineq1}\end{equation}
	
	where the expectations are over the randomness of $\bM$ and of $\pi^*$ and $\pi_{[h]}$. This is Linkage 1.

	To establish Linkage 2, we first focus on a fixed observed history $\hist{\by}$ that has positive probability of being obtained from following $\pi_{[h]}$ (the probability is determined by both the randomness of $\bM$ and the seeding decisions of policy $\pi_{[h]}$). 
	
	By Theorem \ref{thm:greedy_guarantee_step2}, since $\bc(v) = 1 \;\forall v \in \cV$,
	
	\begin{align}
		&\E\left[f(\pit{h}(\bM)@\pi^*(\bM), \bM) - f(\pit{h}(\bM), \bM) \Big| \css{\bM}{\by}, \pit{h}(\bM) = \by\right]\\
		& = \E\left[f(\by@\pi^*(\bM), \bM) - f(\by, \bM) \Big| \css{\bM}{\by}, \pit{h}(\bM) = \by\right]\\ 
		& = \E\left[f(\by@\pi^*(\bM), \bM) - f(\by, \bM) \Big| \css{\bM}{\by}\right]\label{eq:marginal}\\ 
		& \leq \E\left[\mathbf{c}(\pi^*(\bM)) \Big|\css{\bM}{\by}\right] \text{max}_v \left(\Delta(v|\ehist{\by})\right)\\
		& \leq k \cdot \text{max}_v \left(\Delta(v|\ehist{\by})\right). \label{eq:costtok}
	\end{align}
	
	Above, \eqref{eq:marginal} holds because given $\css{\bM}{\by}$, $\pit{h}(\bM) = \by$ is independent of the part of $\bM$ that is outside the coverage of $\hist{\by}$. \eqref{eq:costtok} holds due to the fact that $\pi^*$ is in the form of $\pi'_{[k]}$ for some policy $\pi'$.

	Also, by definition of an $\alpha, \beta$-approximate greedy algorithm, $\pi$ obtains at least $(\beta/\alpha) \text{max}_v \left(\Delta(v|\ehist{\by})\right)$ expected marginal benefit per unit cost in the step immediately following the observation of $\hist{\by}$. Therefore, from \eqref{eq:costtok}, we have
	
	\begin{align}
		&\E\left[f(\pit{h+1}(\bM), \bM) - f(\pit{h}(\bM), \bM) \Big| \css{\bM}{\by}, \pit{h}(\bM) = \by\right] \label{eq:cond1}\\
		&\geq (\beta/\alpha) \text{max}_v \left(\Delta(v|\ehist{\by})\right)\\
		&\geq \frac{\E\left[f(\pit{h}(\bM)@\pi^*(\bM), \bM) - f(\pit{h}(\bM) \Big| \css{\bM}{\by}, \pit{h}(\bM) = \by\right]}{\alpha k/\beta}\label{eq:cond2}.
	\end{align}
	
	We now define $\tilde{p}(\by, \tilde M_{\by}) = \pr\{\css{\bM}{\by}, \pit{h}(\bM) = \by\}$. Then, by \eqref{eq:cond1} - \eqref{eq:cond2}, we have the following:
	
	\begin{align}
		&\E\left[f(\pit{h+1}(\bM), \bM) - f(\pit{h}(\bM), \bM)\right]\\
		&= \sum_{\hist{\by}} \tilde{p}(\by, \tilde M_{\by}) \E\left[f(\pit{h+1}(\bM), \bM) - f(\pit{h}(\bM), \bM) \Big| \css{\bM}{\by}, \pit{h}(\bM) = \by\right]\\
		&\geq \frac{\sum_{\hist{\by}}\tilde{p}(\by, \tilde M_{\by}) \E\left[f(\pit{h}(\bM)@\pi^*(\bM), \bM) - f(\pit{h}(\bM), \bM) \Big| \css{\bM}{\by}, \pit{h}(\bM) = \by\right]}{\alpha k/\beta}\\
		&= \frac{\E\left[f(\pit{h}(\bM)@\pi^*(\bM), \bM)  - f(\pit{h}(\bM), \bM)\right]}{\alpha k/\beta}\label{eq:ineq2}.
	\end{align}
	
	Combining \eqref{eq:ineq1} and \eqref{eq:ineq2}, we get that ror all $h, \; 0 \leq h < l$, \begin{equation}f_{avg}(\pi^*) \leq f_{avg}(\pi_{[h]}) + \frac{\alpha k}{\beta} (f_{avg}(\pi_{[h+1]}) - f_{avg}(\pi_{[h]})).\label{eq:main}\end{equation}
	
	Now define $\Delta_h := f_{avg}(\pi^*) - f_{avg}(\pi_{[h]})$. Then \eqref{eq:main} gives us $\Delta_h \leq \frac{\alpha k}{\beta} (\Delta_h - \Delta_{h+1})$ for all $0 \leq h < l$. Thus, $\Delta_{h+1} \leq (1 - \beta/\alpha k) \Delta_h$ for all $0 \leq h < l$.  Therefore, $\Delta_{l} \leq (1 - \beta/\alpha k)^l \Delta_0$, which by definition of $\Delta_h $, gives us that $f_{avg}(\pi^*) - f_{avg}(\pi) \leq (1 - \beta/\alpha k)^l f_{avg}(\pi^*)$.  By $1-x < e^{-x} \;\;\forall x>0$, we have that $(1 - \beta/\alpha k)^l < e^{-\beta l/\alpha k}$ and thus $f_{avg}(\pi) \geq (1 - e^{-\beta l/\alpha k}) f_{avg}(\pi^*).$ \Halmos
	\endproof
	
	\section{Online Learning Algorithm and Regret Analysis} \label{sct:online}
	
	In this section, we focus on an online-learning setting in which the  influence probabilities are initially unknown. We propose a UCB-based algorithm for our adaptive influence maximization problem with intermediary constraint. Our algorithm needs to learn the influence probabilities while making adaptive seeding decisions to maximize the overall gain under a global budget constraint $T$ with uniform node costs. 
	
	The online-learning version of non-adaptive influence maximization has been explored by \cite{NIPS2017_6895}. In their work, they assume a \textit{linear generalization} of influence probabilities. That is, there exists a vector $\theta \in \mathbb{R}^{d}$ such that the influence probability $\bar{w}(e)$ on arc $e$ is closely approximated by $ \xet \theta$. They then look at a $T$-round influence maximization campaign. In each independent round, the agent selects $K$ seed users and observe the edge semi-bandit feedback of the influence diffusion. The objective is to learn $\theta$ while maximizing the sum of the activated nodes over rounds. Each node can be counted for mulitiple times in their objective. When the rounds are independent, the regret analysis boils down to analyzing the round-wise regrets, and the total regret is the sum of expected regrets over all rounds.

	In our learning problem, we also assumes a linear generalization of edge influence probabilities as in \cite{NIPS2017_6895} as well as the same edge semi-bandit feedback. In our case of adaptive influence maximization, however, the rounds are no longer independent. All previous works incorporating learning and adaptive decision making are epoch-based  \citep{AdaL1, AdaL2}.  Specifically, the time horizon is divided into $n$ \textit{independent epochs}, in each of which the agent makes $T$ rounds of adaptive decisions. Thus, although the problem is adaptive within an epoch, it is not adaptive over the different epochs. Our analysis below is fully adaptive in the sense that there is only one epoch consisting of $T$ rounds of adaptive seeding and parameter learning. Furthermore, under our intermediary constraint, different seeding decisions will result in different edges being removed, and thus causing different diffusion results in the following rounds. Additionally, our objective maximizes the total number of \textit{distinct} nodes that are activated over the course of these $T$ \emph{dependent} rounds. 
	
	The dependency of rounds and the objective of maximizing the sum of reward over \textit{distinct} users lead to two unique challenges. First, we need to evaluate the performance of a sequence of rounds instead of looking at each round individually.  We thus need to interweave the techniques developed in Section \ref{sct:greedy} with techniques to bound the parameter estimation errors during learning. Second, the maximum value of our objective is the sum of rewards of all nodes. Thus $|\cV|$ should be (much) bigger than $T$, since otherwise all nodes will be activated eventually. Therefore, the regret should not have a linear or even square root dependency on $|\cV|$ or $|\cE|$. We pose some assumptions on the graph structure so that it is unlikely for a seed to activate a large number of nodes in any single round. These assumptions will be explained and justified in more details in Section \ref{sct:regret}.  \

	We denote the set of observed edges in each round $t$ as $\obsE{t}$, and the realization of edge $e$ in round $t$ as $y_e^t$. We further assume an access to a $\alpha,\beta$-approximate oracle \ora. Taken as input a graph structure $\cG$, a node reward function $\br$, an influence probability function $w$ and an observed history $\hist{\by}$, $\ora(\cG, \br, w, \hist{\by})$ returns a node $s$ such that $\Delta(s|\ehist{\by}) \geq \frac{1}{\alpha} \max_{u\in \cG}\Delta(u|\ehist{\by})$ with probability at least $\beta$, where $\Delta(\cdot|\cdot)$ is calculated with respect to $\cG$ and $w$. In Appendix \ref{apd:RR}, we give an example of a scalable $\alpha,\beta$-approximation oracle that we extend from the method of Reverse Reachable Sets by \cite{Tang:2014:IMN:2588555.2593670}.
	
	Our UCB-based strategy for the adaptive IM problem, as detailed in Algorithm \ref{alg:1}, updates an upper bound on the influence probability of each edge in every round based on previously observed edge realizations. It then treats the upper bounds as the true influence probabilities and selects a seed for the round using $\ora$ to maximize the marginal increase in the expected number of distinct activated users. The only difference of this learning framework compared to tradiational UCB-based learning algorithms for multi-armed bandit problem (such as the one proposed by \cite{NIPS2017_6895}) is that we need to update the graph $\cG$ at the end of each round to account for edge removals. However, our regret analysis is different due to the dependency among rounds.
	
	\textbf{}
	
	\begin{breakablealgorithm}
		\caption{UCB-based Adaptive IM with Intermediary Constraint (UCB-AIMI)}
		\label{alg:1}
		\begin{algorithmic} 
			\State \textbf{Input}: digraph $\mathcal{G}=\mathcal{(V, E)}$, node rewards $\br: \mathcal{V} \rightarrow \mathbb{R}^+$, edge feature vector $\xe \in \mathbb{R}^d$, budget $T$, hyperparameter $c  \in \dR^+$
			\State \textbf{Initialization}: $\cG' = \cG$, $\bB_0 = 0 \in \mathbb{R}^d$, $\bN_0 = I \in \mathbb{R}^{d\times d}$, $\hist{\by} = [(), [?]^{|\cE| \times T}]$
			\For {$t = 1, ..., T$}
			\State Set $\hth(t-1) \leftarrow \bN_{t-1}^{-1}\bB_{t-1}$
			\State Compute $U_t(e) = Proj_{[0,1]}\left(  \xet\hth(t-1) + c\,\sqrt{\xet\bN_{t-1}^{-1}\xe}\right) $ for all $e$
			\State $v_t \leftarrow$ \ora $(\cG', \br, U_t, \hist{\by})$
			\State Select seed node $v_t$ and observe semi-bandit edge activation realizations
			\State Update $\bN_{t} = \bN_{t-1} + \sum_{e\in \obsE{t}}\xe\xet$, $\bB_t = \bB_{t-1} + \sum_{e\in \obsE{t}}\xe y_e^t$
			\State Update $\cG'$ by removing edges pointing to the intermediary nodes in the round diffusion
			\State Update the observed history $\hist{\by}$
			\EndFor
		\end{algorithmic}
	\end{breakablealgorithm}
	
	\textbf{}

	\subsection{Regret analysis} \label{sct:regret}
	
	\subsubsection*{Assumptions}
	We make the following assumption on the network structure and the influence probabilities. 
	
%
	
	\begin{assumption} \label{asp:1}
	Let $p_{max}$ be the maximum influence probability of an edge. Let $d_{max}$ be the maximum out-degree of a node. We assume that $p_{max} \cdot d_{max} < 1$.\end{assumption}
	
With Assumption \ref{asp:1}, we prove that the number of nodes observed in each round can be upper bounded by any non-negative integer $L$ with probability exponential in $-L$. More concretely,
	\begin{lemma} \label{lemma:3}
		Under Assumption \ref{asp:1}, there exists $R,\rho>0$ such that for any $L \in \dR^+$,
		$$P(\textrm{number of newly activated nodes in each round} > L) \le R e^{-\rho L}.$$
	\end{lemma}

	\proof{Proof.}
	Define a simple branching process as follows:
	\begin{itemize}
	\item $W_0 := 1$;
	\item  $W_n := \text{number of individuals in generation } n = 1, 2, ...$.  In generation $n$, each individual $i = 1, 2, ..., W_n$ produces $\delta_i^n$ offsprings, where $\delta_i^n$'s are independent and identically distributed binomial random variables with parameters $d_{\max}$ and $p_{\max}$.
\end{itemize}

Define $\phi(\lambda) := \ln \E\left(e^{\lambda{\delta_1^0}}\right).$ Let $\phi_n(\lambda) := \phi\left(\lambda + \phi_{n-1}(\lambda)\right)$, for $n = 2, 3, ...,$ and $\phi_1(\lambda) := \phi(\lambda)$. It is easy to see that there exists $\lambda > 0 $ such that $\phi(\lambda)$ is well-defined.

We now show by induction on $n$ that \begin{equation}\label{eq:l5}\ln\E\left(e^{\lambda \sum_{i = 1}^n W_i}\right) = \phi_n(\lambda).\end{equation} 

Consider the base case $n = 1$.  By definition, $E\left(e^{\lambda W_1}\right) =E\left(e^{\lambda \delta_1^0}\right) = \exp\left(\ln\E\left(e^{\lambda \delta_1^0}\right)\right) = \exp(\phi_1(\lambda))$.

Now assume that $\ln\E\left(e^{\lambda \sum_{i = 1}^n W_i}\right) = \phi_n(\lambda)$ holds for $n-1$.  We show that it holds for $n$.  Let $W_{i}^j$ be the number of individuals in generation $i$ in the branching process starting from the $j$th individual in generation $1$, then the left-hand side of the equality can be written in the alternative form below:
\begin{equation} \label{eq:l1}
\begin{split}
\ln\E\left(e^{\lambda \sum_{i = 1}^n W_i}\right) &= \ln \E\left[e^{\lambda \left(\delta_1^0 + \sum_{j = 1}^{\delta_1^0} \sum_{i = 1}^{n-1} W_{i}^{j}\right)}\right]\\
& = \ln \E\left[e^{\lambda \delta_1^0} \left[\E(e^{\lambda \sum_{i = 1}^{n-1} W_i})\right]^{\delta_i^0}\right],
\end{split}
\end{equation}

where the first equality holds because the total number of individuals from generation $1$ to $n$ equals to the sum over the sub-branches each starting with one individual in generation $1$, and the second equality holds because we can take the expectation first conditioned on $\delta_1^0$.

At the same time, the right-hand side of the equality we want to prove can be represented as 
\begin{equation}\label{eq:l2}
\begin{split}
\phi_n(\lambda) &= \phi(\lambda + \phi_{n-1}(\lambda)) \\
& = \ln \E\left[\exp\left({[\lambda + \phi_{n-1}(\lambda)]{\delta_1^0}}\right)\right]\\
& =  \ln \E\left[e^{\lambda \delta_1^0}\exp \left(\phi_{n-1}(\lambda)\right)^{\delta_1^0}\right]\\
& =  \ln \E\left[e^{\lambda \delta_1^0} \left[\E(e^{\lambda \sum_{i = 1}^{n-1} W_i})\right]^{\delta_i^0}\right],
\end{split}
\end{equation}

where the first three equalities are by definition, and the last equality is by our inductive hypothesis.

Equations \eqref{eq:l1} and \eqref{eq:l2} together give us \eqref{eq:l5} as desired.

We now show that there exists an $\lambda'$ such that $0< \lambda' < \lambda$ and $ \phi_{\infty}(\lambda')  := \lim_{n \rightarrow \infty} \phi_n(\lambda') < \infty$. Note that if such $\lambda'$ exists, $\ln\E\left[e^{\lambda'\sum_{i = 1}^\infty W_i}\right] = \phi_{\infty}(\lambda')$ from equality \eqref{eq:l5} we proved above. Then by Markov's inequality, we can get $\E\left[\sum_{i = 1}^\infty W_1 > k\right] = \E\left[\exp(\lambda'\sum_{i = 1}^\infty W_1) > \exp(\lambda' k)\right] \leq \phi_{\infty}(\lambda') e^{-\lambda'k}$ as desired.

To see that such $\lambda'$ exists, we first observe that 
\begin{equation} \label{eq:l3}
\phi_n(0) = \ln\E\left(e^{0 \cdot \sum_{i = 1}^n W_i}\right) = 0 \;\; \forall n.
\end{equation}

Then by chain rule, we have $\phi_n'(\lambda) = \phi'(\lambda + \phi_{n-1}(\lambda))(1 + \phi_{n-1}'(\lambda))$. By \eqref{eq:l3}, we can substitude in $\lambda = 0$ to get $\phi_n'(0) = \phi'(0)(1 + \phi_{n-1}'(0))$. It then easily follows that $\phi_n'(0) = \sum_{i = 1}^n \phi'(0)^i$. Since $\phi'(0) = \E(\delta_1^0) = p_{\max} \cdot d_{\max} < 1$ by assumption, we have that \begin{equation}\label{eq:l4}
\lim_{n\rightarrow \infty}\phi_n'(0) = \lim_{n\rightarrow \infty} \sum_{i = 1}^n \phi'(0)^i < \infty.
\end{equation}

$\lim_{n\rightarrow \infty} \phi_n(0) = 0$ together with \eqref{eq:l4} guarantee that the desired $\lambda'$ exists. 

In each round of our adaptive seeding process, we seed one node. The number of subsequently activated nodes in the diffusion process is stochastically dominated by the size of the entire population of a branching process defined above, starting from the seed node.  	\Halmos

 \endproof

	As we later show, the regret of UCB-AIMI can be upper bounded by an expression that includes the number of new nodes activated in each round. We then use the result of Lemma \ref{lemma:3} to bound this number with high probability. Note that Assumption \ref{asp:1} is sufficient to guarantee that the regret depends on $|\cV|$ and $|\cE|$ at most logarithmically. However, it is not a necessary requirement for it to happen. There can be weaker conditions to ensure it. For instance, as edges are gradually removed over time due to the intermediary constraints, it will be increasingly harder to activate new users in later rounds for any seeding oracles.  Also, we only need the \emph{newly} activated nodes in each round to be relatively small with high probability. The current assumption makes sure that Lemma \ref{lemma:3} holds even  without edge removals and also when counting \textit{all} activated nodes, newly activated as well as the ones that have been activated in previous rounds.  Later, when we conduct numerical experiments, we will generate the test instances without enforcing assumption \ref{asp:1}. We show that our learning algorithm still achieves favorable performance.

	\subsubsection*{Nomenclature} 

	For a summary of nomenclature, please see Table \ref{tab:nomen1}.
	Let $\byo^T = (s_1, s_2, ..., s_T)$ be the seeding sequence selected by UCB-AIMI. Note that $\byo^T$ is a random sequence. For $t = 1, ..., T$, let $\byo^t = (s_1, ..., s_t)$ and use $\tMo^{t}$ to denote the partial matrix at the end of round $t$ after the update. In round $t+1$, UCB-AIMI observes $[\byo^{t}, \tMo^{t}]$ and selects $s_{t+1} = \text{UCB-AIMI}[\byo^{t}, \tMo^{t}]$ as the seed for round $t+1$. Let $\bys^T = (s^*_1, s^*_2, ..., s^*_T)$ be the seeding sequence selected by an optimal policy $*$. Note that $\bys^T$ is also a random sequence.
	
	Use $\Delta_w(s|\ehist{\by})$ to denote the expected increase in reward after seeding node $s$ given any observed history $\hist{\by}$, where the expectation is computed with respect to the influence probabilities $w$. For the true influence probabilities $\bar{w}$, we abbreviate $\Delta_{\bar w}(s|\ehist{\by})$ as $\Delta(s|\ehist{\by})$. We also use $s^g_t$ to denote the seed that the exact greedy algorithm selects in round $t$ given the observed history $[\byo^{t-1}, \tMo^{t-1}]$. That is, $s^g_t \in \argmax_{v\in \cV}\{\Delta(v|\byo^{t-1}, \tMo^{t-1})\}$. 
	
	Let $I_t$ denote the set of nodes that have been activated in rounds $1, ..., t$ following UCB-AIMI. Let $\cE^t_{s,v}$ denotes the set of edges along any path from $s$ to $v$ in $\cG_t$ (the network after edge removals in the first $t-1$ rounds due to intermediary constraints). Use $O_t(e)$ to denote the event that edge $e$'s realization is observed in round $t$ and $\obsE{\tau}$ to denote the set of edges whose realizations are observed in round $\tau$. 
	
	Define the event $\psi_{t-1}$ as $\psi_{t-1} = \{|\xe^\tau \hat{\theta}(\tau -1) - \xe^\tau \theta| \leq c \sqrt{\xe^\tau \bN_{\tau-1}^{-1}\mathbf{x}_e}, \; \forall e \in \mathcal{E}, \tau \leq t\}$. Under $\psi_{t-1}$, $0 \leq U_\tau(e) - \bar{w}(e) \leq 2c\sqrt{\xet N_{\tau-1}^{-1}\xe} \;\;\forall \tau = 1, 2, ..., t$. That is, $\psi_{t-1}$ is the good event, occurring when the estimated upper bounds of the influence probabilities for all edges are not too far away from the true ones for all first $t$ rounds. Later, we will bound the regret under $\psi_{T-1}$ and the probability that its complement $\bar{\psi}_{T-1}$ happens. Note that $\psi_{t-1}$ is measurable with respect to $[\byo^{t-1}, \tMo^{t-1}]$. 
	
	Also define $\xi_{t} = \{|I_\tau \setminus I_{\tau-1}| \leq L \;\; \forall \tau \leq t\}$ to be the event that the number of newly activated nodes is at most $L$ for each of the first $t$ rounds. We will later analyze the regret under $\xi_{T}$ and bound the probability that its complement $\bar{\xi_{T}}$ happens. Lemma \ref{lemma:3} allows us to accomplish the latter. Note that $\xi_{t} $ is measurable with respect to $[\byo^{t}, \tMo^{t}]$.

\begin{center}
	\begin{table}[ht]
		\caption{Nomenclature}
		\label{tab:nomen1}
		\centering
\begin{tabular}{||l|l||}
	
	 \hline 
	\begin{tabular}{c}
		\textbf{Notation}\\
	\end{tabular} &
	\begin{tabular}{l}
		\textbf{Meaning}\\
	\end{tabular}
	\\
	\hline
	\begin{tabular}{l}
		$T$ 
	\end{tabular} &
	\begin{tabular}{l}
		total number of rounds in adaptive seeding
	\end{tabular}
	\\
	\hline 
	\begin{tabular}{l}
	$\byo^t = (s_1, s_2, ..., s_t)$
	\end{tabular} &
	\begin{tabular}{l}
		 random seeding sequence selected by UCB-AIMI in the first \\$t$ rounds
	\end{tabular} \\
	\hline
	
	\begin{tabular}{l}
		$\tMo^{t}$
	\end{tabular} &
	\begin{tabular}{l}
		partial matrix that encodes observed edge realizations after \\ round $t$ following UCB-AIMI     
	\end{tabular}
	\\
	\hline
	
	\begin{tabular}{l}
		UCB-AIMI$[\mathcal{Y}, \tilde{M}]$
	\end{tabular} &
	\begin{tabular}{l}
		seed returned by UCB-AIMI given seeding sequence $\mathcal{Y}$ and\\ partial matrix $\tilde{M}$
	\end{tabular}
	\\
	\hline
	
	\begin{tabular}{l}
		$\Delta_w(s|\ehist{\by})$
	\end{tabular} &
	\begin{tabular}{l}
		expected marginal benefit in reward after seeding $s$ with respect to \\$[\ehist{\by}]$ and influence probabilities $w$
	\end{tabular}
	\\
	\hline
	
	\begin{tabular}{l}
		$\Delta(s|\ehist{\by})$
	\end{tabular} &
	\begin{tabular}{l}
		expected marginal benefit in reward after seeding $s$ with respect to \\$[\ehist{\by}]$ and the true influence probabilities $\bar{w}$
	\end{tabular}
	\\
	\hline
	
	\begin{tabular}{l}
		$s^g_t$
	\end{tabular} &
	\begin{tabular}{l}
		the seed that the exact greedy algorithm selects in round $t$ \\given the observed history $[\byo^{t-1}, \tMo^{t-1}]$, \\i.e.,
		$s^g_t \in \argmax_{v\in \cV}\{\Delta(v|\byo^{t-1}, \tMo^{t-1})\}$
	\end{tabular}
	\\
	\hline
	
	\begin{tabular}{l}
		$I_t$
	\end{tabular} &
	\begin{tabular}{l}
		the set of nodes that have been activated in rounds $1, ..., t$ \\ following UCB-AIMI
	\end{tabular}
	\\
	\hline
	
	\begin{tabular}{l}
		$\cG_t$
	\end{tabular} &
	\begin{tabular}{l}
		the network after edge removals in the first $t-1$ rounds due \\ to intermediary constraints
	\end{tabular}
	\\
	\hline
		
	\begin{tabular}{l}
		$\cE^t_{s,v}$
	\end{tabular} &
	\begin{tabular}{l}
		the set of edges along any path from $s$ to $v$ in $\cG_t$ 
	\end{tabular}
	\\
	\hline
	
	\begin{tabular}{l}
		$O_t(e)$
	\end{tabular} &
	\begin{tabular}{l}
		event that edge $e$'s realization is observed in round $t$ 
	\end{tabular}
	\\
	\hline
	
	\begin{tabular}{l}
		$\obsE{\tau}$
	\end{tabular} &
	\begin{tabular}{l}
		 set of edges whose realizations are observed in round $\tau$. 
	\end{tabular}
	\\
	\hline
	
	\begin{tabular}{l}
		$\psi_{t-1}$
	\end{tabular} &
	\begin{tabular}{l}
		 $\{|\xe^\tau \hat{\theta}(\tau -1) - \xe^\tau \theta| \leq c \sqrt{\xe^\tau \bN_{\tau-1}^{-1}\mathbf{x}_e}, \; \forall e \in \mathcal{E}, \tau \leq t\}$, \\event that the estimated upper bounds of the influence\\ probabilities for all edges are not too far away from the true \\ones for all first $t$ rounds
	\end{tabular}
	\\
	\hline
	
	\begin{tabular}{l}
		$\xi_{t}$
	\end{tabular} &
	\begin{tabular}{l}
		$\{|I_\tau \setminus I_{\tau-1}| \leq L \;\; \forall \tau \leq t\}$, \\event that the number of newly activated nodes is at most \\$L$ for each of the first $t$ rounds.
	\end{tabular}
	\\
	\hline
	
\end{tabular}	
\end{table}

\end{center}

	\subsubsection*{Auxiliary lemmas}
	
	 We detail four auxiliary lemmas before proving the main result.
	
	With $s_t$ being the seed selected by UCB-AIMI in round $t$, Lemma \ref{lemma:M} below bounds the expected marginal benefit of seeding $s_t$  with respect to the UCB of the influence probabilities $U_t$, i.e., $\Delta_{U_t}(s_t|\ehisttto{\by})$, and the true marginal benefit of seeding $s_t$, i.e., $\Delta(s_t|\ehisttto{\by})$, given observed history $[\ehisttto{\by}]$. We later use Lemma \ref{lemma:M} to connect $\Delta(s_t|\ehisttto{\by})$ with $\Delta(s^g_t|\ehisttto{\by})$. The latter is the expected marginal benefit of seeding $s_t^g$, the node selected by the greedy policy under the true influence probabilities given $[\ehisttto{\by}]$, while the former is chosen by an $\alpha,\beta$-approximate greedy under the UCB estimate $U_t$.
	
	Part of the proof extends an analysis in \cite{NIPS2017_6895} so that the terms in the summation over $v$ in $A(t)$ are positive only when $v$ is newly activated in round $t$. This extension allows us to later use Lemma \ref{lemma:3}'s result in bounding $A(t)$.
	\begin{lemma} 
		\label{lemma:M}
		
		Under $\psi_{T-1}$, for any $t = 1, ..., T$,
		\begin{align*}
			&\Delta_{U_t}(s_t|\ehisttto{\by})-\Delta(s_t|\ehisttto{\by}) \le c' \cdot A(t),
		\end{align*}  
		where $c'=\frac{2c}{\min_{e\in \cE} \bar w(e)}$ and $A(t) = \E\left(\sum_{v\in \cV\backslash I_{t-1}}\sum_{e\in \cE^t_{s_t, v}} \mathbf{1}\{O_t(e)\}\mathbf{1}\{v \in I_t\}\sqrt{\xet \bN^{-1}_{t-1}\mathbf{x}_{e}} \; \vert \ehisttto{\by}\right)$.
	\end{lemma}
	
	\proof{Proof.}
	Since $\br \leq 1$, it is sufficient to show that
	\begin{equation}
	\begin{split} \label{eq:a'}
	\Delta_{U_t}&(s_t|\ehisttto{\by})-\Delta(s_t|\ehisttto{\by})\\ 
	& \le c' \cdot \E \left(\sum_{v\in \cV\backslash I_{t-1}}  \br(v) \sum_{e\in \cE^t_{s_t, v}} \mathbf{1}\{O_t(e)\}\mathbf{1}\{v \in I_t\}\sqrt{\mathbf{x}_{e}^t \bN^{-1}_{t-1}\mathbf{x}_{e}}\; \vert \ehisttto{\by}\right).
	\end{split}
	\end{equation}

	To prove \eqref{eq:a'}, we first use $f_t(s_t, w, v)$ to denote the probability that $v$ is activated by $s_t$ given influence probabilities $w$ and observed history $[\ehisttto{\by}]$. Then 	\begin{equation} \label{eq:b'}
	\begin{split}
	\Delta_{U_t}(s_t|\ehisttt{\by})-\Delta(s_t|\ehisttt{\by}) = \sum_{v\in \cV \setminus I_{t-1}} \br(v) \left[f_t(s, U_t, v) - f_t(s_t, \bar w, v)\right].
	\end{split}
	\end{equation}
	
	From Lemma 4 in \citep{NIPS2017_6895}, we know that for any $t$, observed history $[\ehisttto{\by}]$ such that $\psi_{t-1}$ holds, for any node $v \in \cV \setminus \{s_t\}$,
	\begin{equation} \label{eq:a}
	f_t(s_t, U_t, v) - f_t(s_t, \bar w, v) \leq \sum_{e \in \cE^t_{s_t, v}} \frac{\partial{f_t(s_t, \bar{w}, v)}}{\partial{\bar{w}(e)}} [U_t(e) - \bar w(e)].
	\end{equation}
	
	We now modify the proof of Lemma 5 in \citep{NIPS2017_6895} to get an upper bound for $\frac{\partial{f_t(s_t, \bar{w}, v)}}{\partial{\bar{w}(e)}}$.
	In the proof of Lemma 5, \cite{NIPS2017_6895} derived the inequality $$\frac{\partial{f_t(s_t, \bar{w}, v)}}{\partial{\bar{w}(e)}} \leq \E\left[\zeta_1|\ehisttt{\by}, s_t, \bar{w}, \bw(e) = 1\right],$$ where $\zeta_1 := \mathbf{1}\{s_t \text{ influences }v \text{ via at least one path that includes } e\}$. Now use $v_0$ to denote the head node of $e$. Then 
	\begin{equation}
	\begin{split}
	&\E\left[\zeta_1|\ehisttto{\by}, s_t, \bar{w}, \bw(e) = 1 \right]\\ 
	&\leq \E\left[\mathbf{1}\{v_0 \text{ is activated by  } s_t\}\mathbf{1}\{v \text{ is activated by }s_t\}|\ehisttto{\by}, s_t, \bar{w}, \bw(e) = 1\right]\\
	& = \E\left[O_t(e)\mathbf{1}\{v \text{ is activated by }s_t\}|\ehisttto{\by}, s_t, \bar{w}, \bw(e) = 1\right]\\
	& = \E\left[O_t(e)\mathbf{1}\{v \text{ is activated by }s_t\} \textbf{1}\{\bw(e) = 1\}|\ehisttto{\by}, s_t, \bar{w}\right]/\pr(\bw(e) = 1| \bar{w})\\
	& \leq \E\left[O_t(e)\mathbf{1}\{v \text{ is activated by }s_t\}|\ehisttto{\by}, s_t, \bar{w}\right]/\pr(\bw(e) = 1| \bar{w})\\
	& \leq \E\left[O_t(e)\mathbf{1}\{v \text{ is activated by }s_t\}|\ehisttto{\by}, s_t, \bar{w}\right]/\min_{e\in \cE} \bar w(e)\\
	& = \E\left[O_t(e)\mathbf{1}\{v \text{ is activated by }s_t\}|\ehisttto{\by}, \bar{w}\right]/\min_{e\in \cE} \bar w(e),
	\end{split}
	\end{equation}
	where the last equality follows from the fact that $s_t$ is measurable with respect to $\ehisttto{\by}$.
	
	Thus, we have shown that 
	\begin{equation} \label{eq:b}
	\frac{\partial{f_t(s_t, \bar{w}, v)}}{\partial{\bar{w}(e)}} \leq \E\left[\mathbf{1}\{v \text{ is activated by } s_t\}\mathbf{1}\{O_t(e)\}|\ehisttto{\by}\right]/ \min_{e\in \cE} \bar w(e).
	\end{equation}

	Combining \eqref{eq:a} and \eqref{eq:b}, we can conclude that for any $t$, observed history $[\ehisttto{\by}]$ such that $\psi_{t-1}$ holds, for any node $v \in \cV \setminus \{s_t\}$,
	\begin{equation}
	\begin{split}
	&f_t(s_t, U_t, v) - f_t(s_t, \bar w, v) \\ &\leq \frac{1}{\min_{e\in \cE} \bar w(e)}\sum_{e \in \cE^t_{s_t, v}} \E\left(\mathbf{1}\{O_t(e)\}\mathbf{1}\{v \text{ is activated by }s_t\}[U_t(e) - \bar w(e)] | \ehisttto{\by}\right). \label{eq:37}
	\end{split}
	\end{equation}
	
	Since under $\psi_{t-1}$ we have $U_t(e) - \bar w(e) \leq 2c \sqrt{\xe^t \bN_{t-1}^{-1}\mathbf{x}_e}$, from \eqref{eq:37} we further have that 
	\begin{equation} 
	\begin{split}
	&f_t(s_t, U_t, v) - f_t(s_t, \bar w, v)\\ &\leq \frac{2c}{\min_{e\in \cE} \bar w(e)}\sum_{e \in \cE^t_{s_t, v}} \E\left(\mathbf{1}\{O_t(e)\}\mathbf{1}\{v \text{ is activated by }s_t\}\sqrt{\xe^t \bN_{t-1}^{-1}\mathbf{x}_e} | \ehisttto{\by}\right). \label{eq:38}
	\end{split}
	\end{equation}
	
	Inequalities \eqref{eq:b'} and \eqref{eq:38} together give us \eqref{eq:a'}.
	\Halmos
	\endproof
%
Lemma \ref{lemma:k} below lower bounds the expected reward of UCB-AIMI over $T$ rounds, i.e., $\E(f(\byo^T, \bM))$, by a fraction of $\E(f(\bys^T, \bM)))$, the expected optimal reward over $T$ rounds, minus an term that arises due to the UCB estimation errors. The proof uses the results of Theorems \ref{thm:greedy_guarantee_step1} and \ref{thm:greedy_guarantee_step2} as well as that of Lemma \ref{lemma:M}. Lemma \ref{lemma:M} bounds the descrepencies between what the true greedy policy achieves in each step and that achieved by UCB-AIMI in each step due to estimation errors. The proof of Lemma \ref{lemma:k} is essentially incorporating these descrepencies into the proof of Theorem \ref{thm:greedyMain}, the greedy performance guarantee result for the offline problem. Compared to Theorem \ref{thm:greedyMain}, Lemma \ref{lemma:k} has an additional error term.
	\begin{lemma}\label{lemma:k}
		Under $\psi_{T-1}$, $\E(f(\byo^T, \bM)) \ge (1-e^{-\alpha /\beta}) \cdot \E(f(\bys^T, \bM))) - c' \cdot \sum_{t=1}^T \E(A(t))$, where the first two expectations are over the randomness of $\bM$ and the expectation of $A(t)$ is over $\ehisttto{\by}$.
		\end{lemma}
		\proof{Proof.} From Theorem \ref{thm:greedy_guarantee_step1}, we have that under $\psi_{T-1}$, for any $t = 1, 2, ..., T$, 
		\begin{equation}\label{eq:c}
		\E\left[f(\bys^T, \bM) \Big|\csson{t-1}\right] \leq \E\left[f(\byo^{t-1}@\bys^T, \bM) \Big|\csson{t-1}\right].
		\end{equation}
		
		From Theorem \ref{thm:greedy_guarantee_step2}, 
		\begin{equation} \label{eq:d}\E\left[f(\byo^{t-1}@\bys^T, \bM) \Big|\csson{t-1}\right] - f(\ehisttto{\by}) \leq T \cdot \Delta(s_t^g | \ehisttto{\by}).\end{equation}
		
		Inequalities \eqref{eq:c} and \eqref{eq:d} together give us 
		\begin{equation} \label{eq:e}\E\left[f(\bys^T, \bM) \Big|\csson{t-1}\right] - f(\ehisttto{\by}) \leq T \cdot \Delta(s_t^g | \ehisttto{\by}).\end{equation}
		
		Now from Lemma \ref{lemma:M}, we know that under $\psi_{T-1}$, \begin{equation} \label{eq:f}
		\begin{split}
		\Delta(s_t|\ehisttto{\by}) &\geq \Delta_{U_t}(s_t|\ehisttto{\by})- c' \cdot A(t)\\
		& \geq \frac{\beta}{\alpha} \Delta_{U_t}(s^g_t|\ehisttto{\by})- c' \cdot A(t)\\
		& \geq \frac{\beta}{\alpha} \Delta(s^g_t|\ehisttto{\by})- c' \cdot A(t),
		\end{split}
		\end{equation}
		where the second inequality follows from the fact that $s_t$ is chosen by an $\alpha,\beta$-approximate greedy oracle under $U_t$, and the last inequality follows from the fact that under $\psi_{T-1}$, $U_t \geq \bar w$.
		
		Inequalities \eqref{eq:e} and \eqref{eq:f} together show that under $\psi_{T-1}$, 
		\begin{equation} \label{eq:g}\E\left[f(\bys^T, \bM) \Big|\csson{t-1}) - f(\ehisttto{\by}\right] \leq T \cdot \frac{\alpha}{\beta}\left(\Delta(s_t | \ehisttto{\by}) + c' \cdot A(t)\right).
		\end{equation}
		
		Now define $\beta_{t} := \E\left[f(\bys^{T}, \bM) \Big| \csson{t}\right] - f(\ehistto{\by})$.  Then we can write \eqref{eq:g} as 
		
		\begin{equation} \label{eq:h} \frac{\beta_{t-1}}{T} \leq \frac{\alpha}{\beta}\left(c' \cdot A(t) + \beta_{t-1} - \E(\beta_t| \ehisttto{\by})\right).
		\end{equation}
		
		Since \eqref{eq:h} holds for all $t = 1, ..., T$ under $\psi_{T-1}$, and $1-\frac{\alpha}{\beta \cdot T} \leq 1$, we have that 
		\begin{equation} 
		\begin{split}
		\label{eq:i} \E(\beta_{T}) &\leq \left(1-\frac{\alpha}{\beta \cdot T}\right)^T \cdot \E(\beta_0) + c' \cdot \sum_{t = 1}^T \E(A(t))\\
		&\leq e^{-\alpha/\beta} \cdot \E(\beta_0) + c' \cdot \sum_{t = 1}^T \E(A(t)). \Halmos
		\end{split}
		\end{equation}
		
		\endproof

Lemma \ref{lemma:A} below states that for any $\delta \in (0,1)$, when $c$ exceeds a value determined by $\delta$ and some other problem instance specific parameters, the probability that good event does not happen is at most $\gamma$. Recall that the good event is defined as $\psi_{t-1} := \{|\xe^\tau \hat{\theta}(\tau -1) - \xe^\tau \theta| \leq c \sqrt{\xe^\tau \bN_{\tau-1}^{-1}\mathbf{x}_e}, \; \forall e \in \mathcal{E}, \tau \leq t\}$. Under $\psi_{t-1}$, $0 \leq U_\tau(e) - \bar{w}(e) \leq 2c\sqrt{\xet N_{\tau-1}^{-1}\xe} \;\;\forall \tau = 1, 2, ..., t$.
	\begin{lemma}[\cite{NIPS2017_6895}, Lemma 2] \label{lemma:A}
		For any $t = 1, 2, ..., T$ and any $\delta \in (0,1)$, if $$c \geq \sqrt{d \log (1+T\cdot |\cE|/d) + 2 \log (1/\delta)} + ||\theta||,$$ then $\pr(\bar{\psi}_{t-1}) \leq \delta$.
		\end{lemma}

In Lemma \ref{lemma:B}, we bound the error term appeared in Lemma \ref{lemma:k} conditioned on the good event $\psi_{T-1}$ and the other event $ \xi_{T}$ or its complement we defined at the beginning of the section. Recall that $\xi_{T}$ is the event that in all $T$ rounds, the newly activated nodes of each round does not exceed $L$. Given $\xi_{T}$, we can bound each $A(t)$ by replacing $\sum_{v\in \cV\backslash I_{t-1}} \mathbf{1}\{v \in I_t\}$ with $L$. Given $\bar{\xi}_{T}$ on the other hand, $\sum_{v\in \cV\backslash I_{t-1}}\mathbf{1}\{v \in I_t\}$ can only be upper bounded by $|\cV|$. Besides these observations, we also uses an intermediary result in the proof of Lemma 1 by \cite{NIPS2017_6895}.

Later, we can use the law of total probability to express $\sum_{t=1}^T \E(A(t))$ as a weighted sum of $\sum_{t=1}^T \E(A(t)| \psi_{T-1}, \xi_{T})$ and $\sum_{t=1}^T \E(A(t)| \psi_{T-1}, \bar{\xi}_{T})$ and some other terms. Then with Lemma \ref{lemma:B}, we can upper bound $\sum_{t=1}^T \E(A(t))$.
	\begin{lemma}\label{lemma:B}
		Let $l_{max}$ be the maximum outdegree in $\cG$.  Then
		\begin{equation*}
		\begin{split}
		&\sum_{t=1}^T \E(A(t)| \psi_{T-1}, \xi_{T}) \leq L^2 \cdot l_{max} \cdot \sqrt{Td \cdot \log(1 + T L \cdot l_{max} /d)},\quad \mbox{and}\\
		&\sum_{t=1}^T \E(A(t)| \psi_{T-1}, \bar{\xi}_{T}) \leq |\cV| \cdot |\cE| \cdot \sqrt{Td \cdot \log(1 + T |\cE|/d)}.
		\end{split}
		\end{equation*}
	\end{lemma}
	\proof{Proof.} 
	First, recall that $$A(t) = \E\left(\sum_{v\in \cV\backslash I_{t-1}}\sum_{e\in \cE^t_{s_t, v}} \mathbf{1}\{O_t(e)\}\mathbf{1}\{v \in I_t\}\sqrt{\xet \bN^{-1}_{t-1}\mathbf{x}_{e}} \; \vert \ehisttto{\by}\right).$$
	
	Taking expectation of each $A(t)$ over $\ehisttto{\by}$ such that $\psi_{T-1}$ holds, we obtain $$\E(A(t)| \psi_{T-1}) = \E\left(\sum_{v\in \cV\backslash I_{t-1}}\sum_{e\in \cE^t_{s_t, v}} \mathbf{1}\{O_t(e)\}\mathbf{1}\{v \in I_t\}\sqrt{\xet \bN^{-1}_{t-1}\mathbf{x}_{e}} \; \vert \psi_{T-1}\right),$$ and thus
	\begin{equation}
	\begin{split}
	\sum_{t=1}^T \E(A(t)| \psi_{T-1}) &= \E\left(\sum_{t=1}^T \sum_{v\in \cV\backslash I_{t-1}}\sum_{e\in \cE^t_{s_t, v}} \mathbf{1}\{O_t(e)\}\mathbf{1}\{v \in I_t\}\sqrt{\xet \bN^{-1}_{t-1}\mathbf{x}_{e}} \; \Big \vert \psi_{T-1}\right)\\
	& = \E\left(\sum_{t=1}^T \sum_{e\in \cE_t^o} \sqrt{\xet \bN^{-1}_{t-1}\mathbf{x}_{e}} \sum_{v\in \cV\backslash I_{t-1}} \mathbf{1}\{v \in I_t\} \mathbf{1}\{e\in \cE^t_{s_t, v}\} \; \Big \vert \psi_{T-1}\right).
	\end{split}
	\end{equation}
	
	In the proof of Lemma 1 in \cite{NIPS2017_6895}, the authors show that if for all rounds $t$, $m$ is an upper bound on $|\cE_t^o|$ (the number of observed edge realizations in round $t$), then we have
		\begin{equation} \label{eq:x}
		\sum_{t=1}^T \sum_{e\in \cE_t^o} \xet \bN^{-1}_{t-1}\mathbf{x}_{e} \leq d m \cdot \log(1 + T m/d).
		\end{equation} 
		Applying Cauchy-Schwartz Inequality with \eqref{eq:x}, we have that 
		
 		\begin{equation*}
			\begin{split}
			\sum_{t=1}^T \sum_{e\in \cE_t^o} \sqrt{\xet \bN^{-1}_{t-1}\mathbf{x}_{e}} &\leq \sqrt{\sum_{t=1}^T \sum_{e\in \cE_t^o} 1  \cdot \sum_{t=1}^T \sum_{e\in \cE_t^o} \xet \bN^{-1}_{t-1}\mathbf{x}_{e}}\\
			&\leq \sqrt{mT \cdot d m \cdot \log(1 + T m/d)}\\ 
			&\leq m \cdot \sqrt{Td \cdot \log(1 + T m/d)}.
			\end{split}
			\end{equation*} 
	
	We now consider $\sum_{t=1}^T \E(A(t)| \psi_{T-1}, \xi_{T})$. Under $\xi_{T}$, the number of nodes activated in each round is at most $L$, and thus $\sum_{v\in \cV\backslash I_{t-1}} \mathbf{1}\{v \in I_t\} \leq L$. Also, since $l_{max}$ is the maximum outdegree of $\cG$, we know that under $\xi_{T}$, $|\cE_t^o| \leq L\cdot l_{max}$.  Therefore, 
	\begin{equation}
	\begin{split}
	\sum_{t=1}^T \E(A(t)| \psi_{T-1}, \xi_{T}) 
	& = \E\left(\sum_{t=1}^T \sum_{e\in \cE_t^o} \sqrt{\xet \bN^{-1}_{t-1}\mathbf{x}_{e}} \sum_{v\in \cV\backslash I_{t-1}} \mathbf{1}\{v \in I_t\} \mathbf{1}\{e\in \cE^t_{s_t, v}\} \; \Big \vert \psi_{T-1}, \xi_T\right)\\
	& \leq \E\left(\sum_{t=1}^T \sum_{e\in \cE_t^o} \sqrt{\xet \bN^{-1}_{t-1}\mathbf{x}_{e}} \sum_{v\in \cV\backslash I_{t-1}} \mathbf{1}\{v \in I_t\} \; \Big \vert \psi_{T-1}, \xi_T\right)\\
	& \leq L \cdot \E\left(\sum_{t=1}^T \sum_{e\in \cE_t^o} \sqrt{\xet \bN^{-1}_{t-1}\mathbf{x}_{e}}  \; \Big \vert \psi_{T-1}, \xi_T\right)\\
	& \leq L^2 \cdot l_{max} \cdot \sqrt{Td \cdot \log(1 + T L \cdot l_{max} /d)}.
	\end{split}
	\end{equation}
	
	On the other hand, under $\bar{\xi}_{T}$, the number of nodes activated in each round is at most $|\cV|$, and thus $\sum_{v\in \cV\backslash I_{t-1}} \mathbf{1}\{v \in I_t\} \leq |\cV|$. Also, $|\cE_t^o| \leq |\cE|$. Therefore, 
	\begin{equation}
	\begin{split}
	\sum_{t=1}^T \E(A(t)| \psi_{T-1}, \bar{\xi}_{T}) 
	& = \E\left(\sum_{t=1}^T \sum_{e\in \cE_t^o} \sqrt{\xet \bN^{-1}_{t-1}\mathbf{x}_{e}} \sum_{v\in \cV\backslash I_{t-1}} \mathbf{1}\{v \in I_t\} \mathbf{1}\{e\in \cE^t_{s_t, v}\} \; \Big \vert \psi_{T-1}, \bar{\xi}_T\right)\\
	& \leq \E\left(\sum_{t=1}^T \sum_{e\in \cE_t^o} \sqrt{\xet \bN^{-1}_{t-1}\mathbf{x}_{e}} \sum_{v\in \cV\backslash I_{t-1}} \mathbf{1}\{v \in I_t\} \; \Big \vert \psi_{T-1}, \bar{\xi}_T\right)\\
	& \leq |\cV| \cdot \E\left(\sum_{t=1}^T \sum_{e\in \cE_t^o} \sqrt{\xet \bN^{-1}_{t-1}\mathbf{x}_{e}}  \; \Big \vert \psi_{T-1}, \bar{\xi}_T\right)\\
	& \leq |\cV| \cdot |\cE| \cdot \sqrt{Td \cdot \log(1 + T |\cE|/d)}.	\Halmos
	\end{split}
	\end{equation}
	\endproof
	
	\subsubsection*{Major result}

	The regret of our UCB-AIMI algorithm is summarized in the following theorem.
	
	The results shows that compared to the greedy performance guarantee of the offline problem in Theorem \ref{thm:greedyMain}, learning the parameters introduces an additional error term that is in the order of square root of $T$ if we ignore all the logorithmic factors. Note that this error term depends only logorithmically on $|\cE|$ and $|\cV|$ as desired. Also, since each node's reward is lower bounded by $r$, $\E(f(\bys^T, \bM))$ grows at least linearly in $T$. Thus the result of a square root dependency of the error term on $T$ is nontrivial.

	\begin{theorem} \label{thm:1}
	Let $\tilde{\mathcal{O}}$ denote order $O()$ notation that that accounts only for terms of order higher than logarithmic, and $l_{max}$ be the maximum outdegree in $\cG$. Assume that Assumption \ref{asp:1} holds. Recall from Lemma \ref{lemma:3}, there exists $R, \rho >0$ such that $$P(\textrm{number of newly activated nodes in each round} > L) \le R e^{-\rho L}.$$ When the intermediary constraint, we have that 
	\begin{equation*}
		\begin{split}
		\E(f(\byo^T, \bM))
		&\geq (1-e^{-\alpha /\beta}) \cdot \E(f(\bys^T, \bM))\\
		&  \;\;\;\; - \mathcal{O} \biggl( \sqrt{T} \biggl[\frac{l_{max}^{3/2}}{\rho^{5/2}\cdot \min_{e\in \cE} \bar w(e)} \cdot \left(\sqrt{d \log(T |\cE|/d) + \log |\cV|} + ||\theta||\right) \\
		& \;\;\;\;\; \;\;\;\;\; \;\;\;\;\; \;\;\;\; \;\;\;\;\; \;\;\; \cdot \log^2 (|\cV||\cE|T) \sqrt{\log(T \log(|\cV||\cE|T))} + \sqrt{d} R \sqrt{\log(T |\cE|/d)}\biggr ]\biggr )\\
		& = (1-e^{-\alpha /\beta}) \cdot \E(f(\bys^T, \bM)) - \tilde{\mathcal{O}} \left(  \sqrt{T} \cdot \left [\frac{l_{max}^{3/2}}{\rho^{5/2}\cdot \min_{e\in \cE} \bar w(e)}(d + ||\theta||) + \sqrt{d}R \right] \right).
		\end{split}
	\end{equation*}
	
 	Note that since $\br(v) \geq r \;\; \forall v \in \cV$, $\E(f(\bys^T, \bM))$ must grow at least linearly in $T$. 
	\end{theorem}

	\proof{Proof.}
	
	We first use the law of total probability to break $\E(f(\byo^T, \bM))$ into two parts. Recall that $\E(f(\byo^T, \bM))$ is the expected reward of following the sequence $\byo^T$ chosen by our algorithm UCB-AIMI. 
	\begin{equation}\label{eq:52}
	\begin{split}
		\E(f(\byo^T, \bM)) &= \E(f(\byo^T, \bM)| \psi_{T-1}) \pr(\psi_{T-1}) + \E(f(\byo^T, \bM)| \bar{\psi}_{T-1}) \pr(\bar{\psi}_{T-1})\\
		& \geq \E(f(\byo^T, \bM)| \psi_{T-1}) \pr(\psi_{T-1}). 
	\end{split}
	\end{equation}

	Note that \eqref{eq:52} holds because $f(\byo^T, \bM)$ and $\pr(\bar{\psi}_{T-1}) \geq 0$ are always non-negative. 
	
	Next, we use Lemma \ref{lemma:k} to express $\E(f(\byo^T, \bM)| \psi_{T-1})$ in terms of $\E(f(\bys^T, \bM))$ and use the law of total probability again to break $\E(f(\bys^T, \bM))$ into a weighted sum of conditional probabilities:
	\begin{align*}
		 \E(f(\byo^T, \bM)| \psi_{T-1}) \pr(\psi_{T-1})
		& \geq (1-e^{-\alpha /\beta}) \cdot \E(f(\bys^T, \bM) | \psi_{T-1}) \pr(\psi_{T-1}) - c' \cdot \sum_{t=1}^T \E(A(t) |\psi_{T-1})\pr(\psi_{T-1}) \\
		& = (1-e^{-\alpha /\beta}) \cdot \E(f(\bys^T, \bM)) - (1-e^{-\alpha /\beta}) \cdot \E(f(\bys^T, \bM) | \bar{\psi}_{T-1}) \pr(\bar{\psi}_{T-1})\\ 
		& \;\;\;\;\;\;- c' \cdot \sum_{t=1}^T \E(A(t) |\psi_{T-1})\pr(\psi_{T-1}).
	\end{align*}

	Now, since $\E(f(\bys^T, \bM) | \bar{\psi}_{T-1}) \leq |\cV|$ and $\pr(\xi_T | \psi_{T-1}) \leq 1$, we have that 

	\begin{equation} \label{eq:m'}
	\begin{split}
	&(1-e^{-\alpha /\beta}) \cdot \E(f(\bys^T, \bM)) - (1-e^{-\alpha /\beta}) \cdot \E(f(\bys^T, \bM) | \bar{\psi}_{T-1}) \pr(\bar{\psi}_{T-1})\\ 
	& \;\;\;\;\;\;- c' \cdot \sum_{t=1}^T \E(A(t) |\psi_{T-1})\pr(\psi_{T-1}).\\
	& \geq (1-e^{-\alpha /\beta}) \cdot \E(f(\bys^T, \bM)) - (1-e^{-\alpha /\beta}) \cdot |\cV| \cdot \pr(\bar{\psi}_{T-1}) \\
	&\;\;\;\;\;\; - c' \cdot \left(\sum_{t=1}^T \E(A(t) |\psi_{T-1}, \xi_{T}) \pr(\xi_{T} | \psi_{T-1}) + \sum_{t=1}^T \E(A(t) |\psi_{T-1}, \bar{\xi}_{T}) \pr(\bar{\xi}_{T} | \psi_{T-1})\right)\\
	& \geq (1-e^{-\alpha /\beta}) \cdot \E(f(\bys^T, \bM)) - (1-e^{-\alpha /\beta}) \cdot |\cV| \cdot \pr(\bar{\psi}_{T-1})\\
	&\;\;\;\;\;\; - c' \cdot \left(\sum_{t=1}^T \E(A(t) |\psi_{T-1}, \xi_{T}) + \sum_{t=1}^T \E(A(t) |\psi_{T-1}, \bar{\xi}_{T}) \pr(\bar{\xi}_{T} | \psi_{T-1})\right). 
	\end{split}
\end{equation}
	
		From Lemma \ref{lemma:A}, if \begin{equation} \label{eq:m} c = \sqrt{d \log (1+T\cdot |\cE|/d) + 2 \log (|\cV|)} + ||\theta||,\end{equation} then $\pr(\bar{\psi}_{T-1}) \leq 1/|\cV|$. Now by Lemma \ref{lemma:3}, $\pr(\bar{\xi}_{T}) \leq TRe^{-\rho L}$.  Therefore, $$\pr(\bar{\xi}_{T}| \psi_{T-1}) = \frac{\pr(\bar{\xi}_{T}, \psi_{T-1})}{ 1- \pr(\bar{\psi}_{T-1})} \leq \frac{|\cV|\cdot \pr(\bar{\xi}_{T})}{ |\cV|-1} \leq \frac{|\cV|}{ |\cV|-1} TRe^{-\rho L}.$$
		
		Now set $L = \log(|\cV|\cdot |\cE|\cdot T)/\rho$. Then  
		\begin{equation} \label{eq:l}
		\pr(\bar{\xi}_{T}| \psi_{T-1}) \leq \frac{|\cV|}{ |\cV|-1} TRe^{-\rho L} = \frac{R}{  |\cE|(|\cV|-1)}.
		\end{equation}
		
		Also, from Lemma \ref{lemma:B}, \begin{equation} \label{eq:y}
			\begin{split}
				\sum_{t=1}^T \E(A(t)| \psi_{T-1}, \xi_{T}) &\leq L^2 \cdot l_{max} \cdot \sqrt{Td \cdot \log(1 + T L \cdot l_{max} /d)},\\
				& \leq \frac{\log^2(|\cV|\cdot |\cE|\cdot T)\cdot l_{max}}{\rho^2}\sqrt{Td \cdot \log(1 + T \log(|\cV|\cdot |\cE|\cdot T) \cdot l_{max} /d\rho)}.
				\end{split}
			\end{equation}
			
		Lemma \ref{lemma:B} together with \eqref{eq:l} give us
		\begin{equation} \label{eq:z}
		\begin{split}
				&\sum_{t=1}^T \E(A(t)| \psi_{T-1}, \bar{\xi}_{T}) \pr(\bar{\xi}_{T}| \psi_{T-1}) \leq \frac{R |\cV|}{|\cV|-1} \sqrt{Td \cdot \log(1 + T |\cE|/d)} .
			\end{split}
		\end{equation}
		
		Combining \eqref{eq:m'}, \eqref{eq:m}, \eqref{eq:y} and \eqref{eq:z}, we have that
		\begin{equation*}
		\begin{split}
			\E(f(\byo^T, \bM)) \geq 
			& (1-e^{-\alpha /\beta}) \cdot \E(f(\bys^T, \bM)) - (1-e^{-\alpha /\beta}) \cdot |\cV| \cdot \pr(\bar{\psi}_{T-1})\\
			&\;\;\;\;\;\; - c' \cdot \left(\sum_{t=1}^T \E(A(t) |\psi_{T-1}, \xi_{T}) + \sum_{t=1}^T \E(A(t) |\psi_{T-1}, \bar{\xi}_{T}) \pr(\bar{\xi}_{T} | \psi_{T-1})\right)\\
			& \geq (1-e^{-\alpha /\beta}) \cdot \E(f(\bys^T, \bM)) - (1-e^{-\alpha /\beta})\\
			&\;\;\;\;\;\; - \frac{2c}{\min_{e\in \cE} \bar w(e)} \cdot \left(\sum_{t=1}^T \E(A(t) |\psi_{T-1}, \xi_{T}) + \sum_{t=1}^T \E(A(t) |\psi_{T-1}, \bar{\xi}_{T}) \pr(\bar{\xi}_{T} | \psi_{T-1})\right)\\
			& \geq (1-e^{-\alpha /\beta}) \cdot \E(f(\bys^T, \bM)) - (1-e^{-\alpha /\beta})\\
			&\;\;\;\;\;\; - \frac{2	(\sqrt{d \log (1+T\cdot |\cE|/d) + 2 \log (|\cV|)} + ||\theta||)}{\min_{e\in \cE} \bar w(e)} \\
			&\;\;\;\;\;\;  \cdot \Big(\frac{\log^2(|\cV|\cdot |\cE|\cdot T)\cdot l_{max}}{\rho^2}\sqrt{Td \cdot \log(1 + T \log(|\cV|\cdot |\cE|\cdot T) \cdot l_{max} /d\rho)} \\
			& \;\;\;\;\;\;\;\;\;\; + \frac{R |\cV|}{|\cV|-1} \sqrt{Td \cdot \log(1 + T |\cE|/d)}\Big) \\
			& = (1-e^{-\alpha /\beta}) \cdot \E(f(\bys^T, \bM))\\
			&  \;\;\;\; - \mathcal{O} \biggl( \sqrt{T} \biggl[\frac{l_{max}^{3/2}}{\rho^{5/2}\cdot \min_{e\in \cE} \bar w(e)} \cdot \left(\sqrt{d \log(T |\cE|/d) + \log |\cV|} + ||\theta||\right) \\
			& \;\;\;\;\; \;\;\;\;\; \;\;\;\;\; \;\;\;\; \;\;\;\;\; \;\;\; \cdot \log^2 (|\cV||\cE|T) \sqrt{\log(T \log(|\cV||\cE|T))} + \sqrt{d} R \sqrt{\log(T |\cE|/d)}\biggr ]\biggr )\\
			& = (1-e^{-\alpha /\beta}) \cdot \E(f(\bys^T, \bM)) - \tilde{\mathcal{O}} \left(  \sqrt{T} \cdot \left [\frac{l_{max}^{3/2}}{\rho^{5/2}\cdot \min_{e\in \cE} \bar w(e)}(d + ||\theta||) + \sqrt{d}R \right] \right). \Halmos
			\end{split}
			\end{equation*}
	\endproof

	\section{Numerical Experiments}
	
	To test our adaptive seeding and learning algorithms, we use four Twitter subnetworks from \cite{SNAP} to conduct numerical experiments. In these experiments, we compare the performance of five seeding oracles. They are:
	
	\begin{enumerate}
		\item[(a)] {\tt{rdm}} - a random seeding oracle which uniformly selects a node from the network at random as the seed in each round (we use its performance as a baseline),
		\item[(b)] {\tt{bgg\_dgr}} - a seeding oracle that in each round seeds a node with the biggest out-degree in the updated network (some empirical studies have shown favorable performance of this seeding strategy, see \cite{bdg_emp}),
		\item[(c)] {\tt{grd\_kw}} - an approximate greedy seeding oracle that knows the true influence probabilities on the edges (we use it as a proxy for the optimal seeding oracle, since Theorem \ref{thm:greedyMain} guarantees that it achieves at least a constant fraction of the optimal oracle reward),
		\item[(d)] {\tt{grd\_lf}} - an approximate greedy seeding oracle that does not know the true influence probabilities on the edges, but assumes linear generalization of influence probabilities (i.e., our proposed UCB-AIMI algorithm),
		\item[(e)] {\tt{grd\_lnf}} - an approximate greedy seeding oracle that does not know the true influence probabilities on the edges, and does not assume linear generalization of influence probabilities (i.e. the CUCB algorithm proposed in \cite{CUCB}).
		\end{enumerate}
	 
	 To construct \textit{edge feature vectors} in $\mathbb{R}^d$, we first use the {\tt{node2vec}} algorithm proposed by \cite{DBLP:journals/corr/GroverL16} to construct \textit{node feature vectors} in $\mathbb{R}^d$. Then for each edge, we use the element-wise product of its head and tail nodes' node feature vectors as its edge feature vector. Since influence probabilities on the edges are not known, we randomly sample an influence probability from {\tt{Unif}}(0,0.1) for each edge. We adopt these setups from \cite{NIPS2017_6895}.  Clearly, it is very unlikely that there exists a vector $\theta$ that perfectly generalizes the influence probabilities. Nevertheless, we compare our feature-based learning oracle {\tt{grd\_lf}} with the non-feature based one {\tt{grd\_lnf}}. We treat these edge feature vectors and influence probabilities as ground truth in the experiments. We set the number of rounds as half of the number of nodes and $d = 5$ for all the experiments we conduct. 
	 
	 The four sub-networks we chose have different edge densities. The first network has 67 nodes and 607 directed edges. A complete directed graph with 67 nodes has 4422 edges, and around 14\% of these edges are present in our first sub-network. The second sub-network has 138 nodes and 719 dircted edges. Only around 4\% possible directed edges are present. The third sub-network is the densist, with 145 nodes and 4146 directed edges. This corresponds to having 20\% of possible edges present. The fourth sub-network has 243 nodes and 6561 directed edges, i.e., 11\% possible edges present.  
	 
\begin{figure}[h]
	\centering
	\includegraphics[width=10cm]{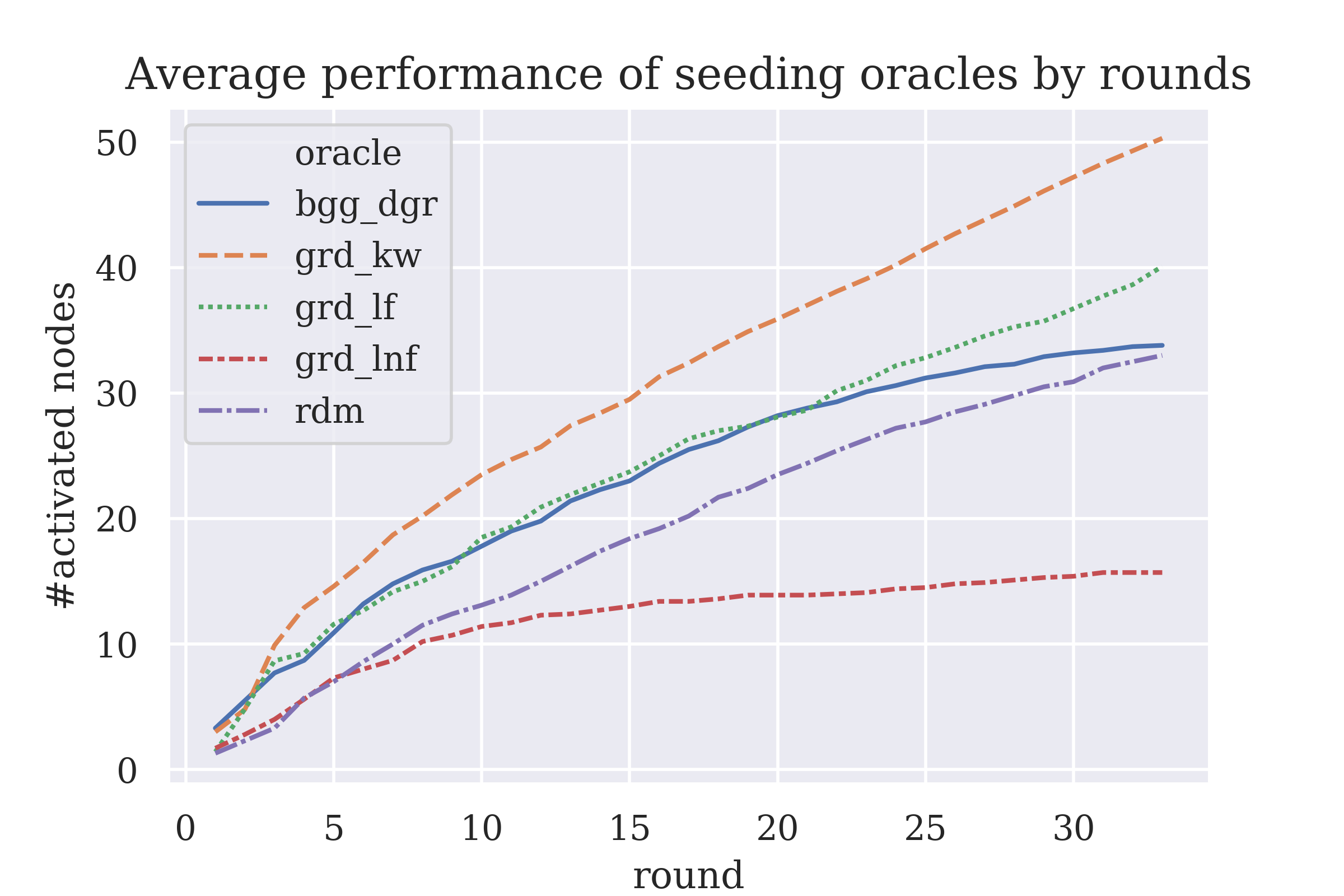}
	\caption{Performance of five seeding oracles by rounds, averaged over 10 realizations, on a Twitter network with 67 nodes and 507 directed edges.}
	\label{fig:477094958}
\end{figure}

\begin{figure}[h]
	\centering
	\includegraphics[width=10cm]{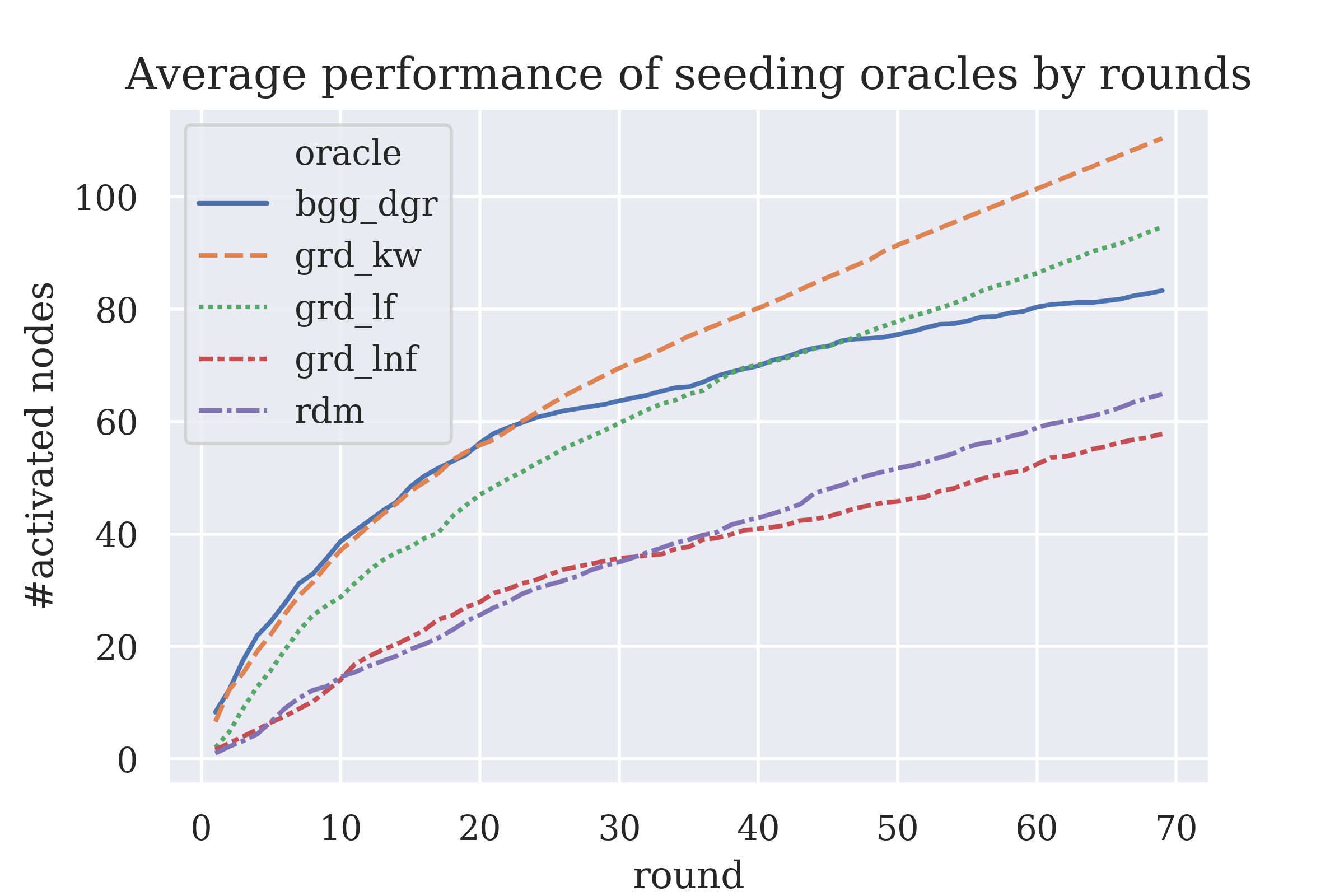}
	\caption{Performance of five seeding oracles by rounds, averaged over 10 realizations, on a Twitter network with 138 nodes and 719 directed edges.}
	\label{fig:441252694}
\end{figure}

\begin{figure}[h]
	\centering
	\includegraphics[width=10cm]{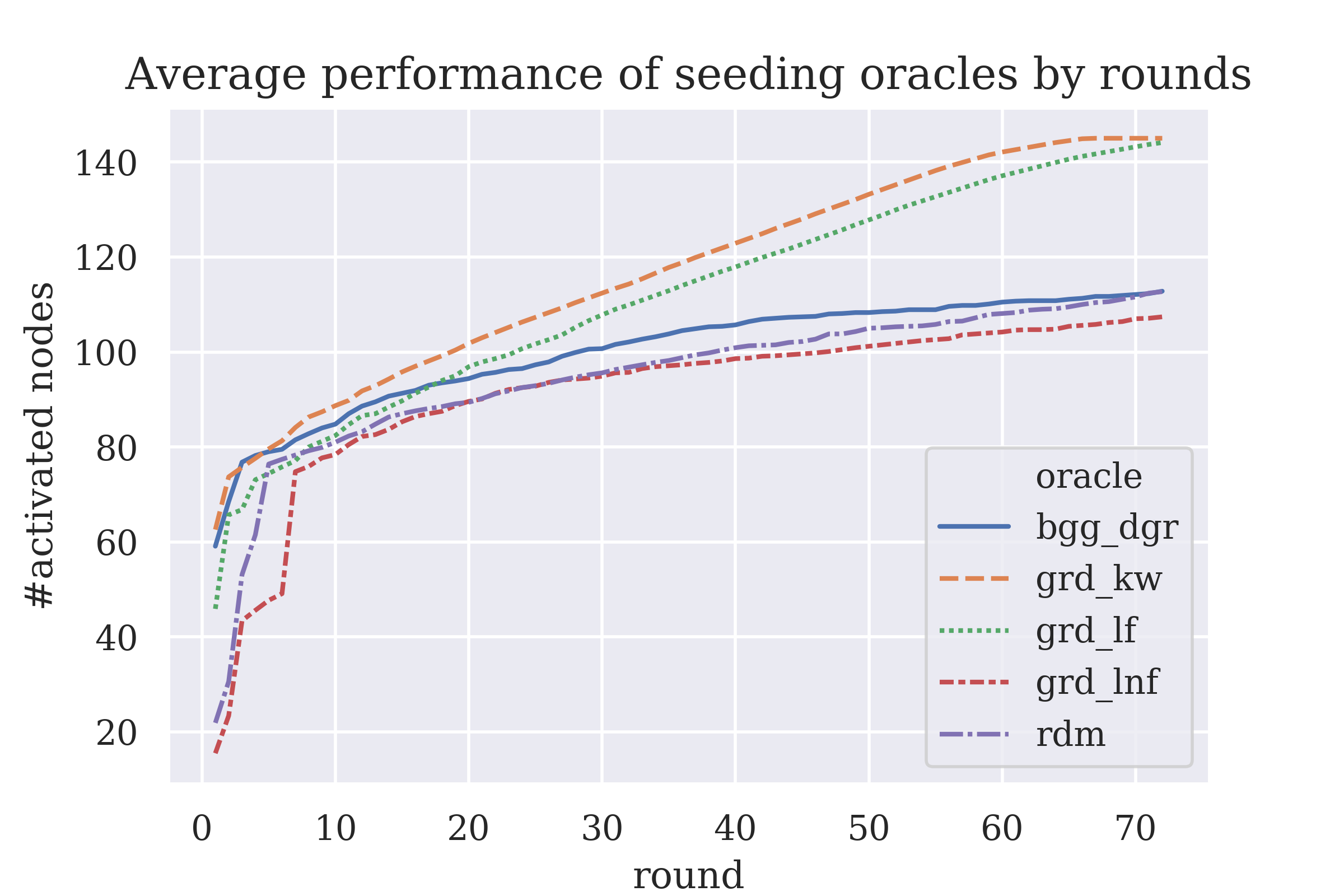}
	\caption{Performance of five seeding oracles by rounds, averaged over 10 realizations, on a Twitter network with 145 nodes and 4145 directed edges.}
	\label{fig:434433610}
\end{figure}

\begin{figure}[h]
	\centering
	\includegraphics[width=10cm]{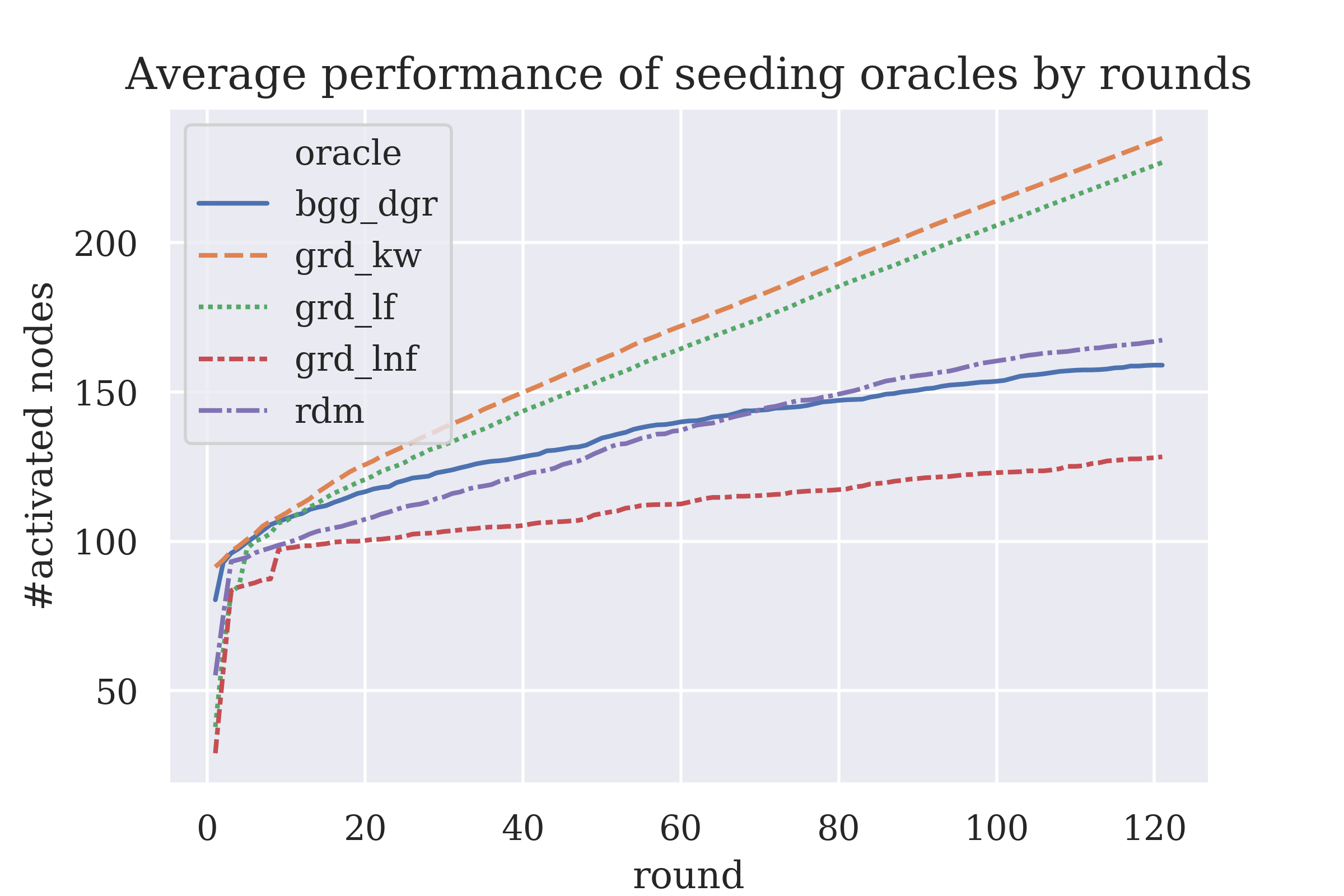}
	\caption{Performance of five seeding oracles by rounds, averaged over 10 realizations, on a Twitter network with 243 nodes and 6561 directed edges.}
	\label{fig:745823}
\end{figure}

	 As we can see from Figure \ref{fig:477094958}, \ref{fig:441252694}, \ref{fig:434433610} and \ref{fig:745823}, despite the imperfect linear generalization of edge influence probabilities, in all four networks with different densities, {\tt{grd\_lf}} consistently outperforms {\tt{grd\_lnf}} and {\tt{rdm}}. In the three denser networks (Figure \ref{fig:477094958}, \ref{fig:434433610}, and \ref{fig:745823}), {\tt{grd\_lf}} also performs on par or better than {\tt{bgg\_dgr}}, especially after the first 10 rounds. Even in the sparse network (Figure \ref{fig:441252694}), {\tt{grd\_lf}} exceeds {\tt{bgg\_dgr}} by an increasing margin after the 45th round. The sub-par performance of {\tt{grd\_lnf}} might be due to the fact that the number of rounds in our adaptive seeding setting is too small for it to learn a good approximation of edge influence probabilities. Since CUCB initiates all edge probabilities to be 1's, nodes with unobserved out-edges might be preferred. The oracle is very likely still mostly ``exploring'' before the adaptive seeding process ends.  
	 
	 The two largest networks (Figure \ref{fig:434433610} and \ref{fig:745823}) give rise to an interesting and different pattern: the average number of activated nodes after the first round exceeds 40 for three oracles in both networks,  {\tt{bgg\_dgr}}, {\tt{grd\_kw}}, and {\tt{grd\_lf}}. For the remaining two oracles,  {\tt{grd\_lnf}} and {\tt{rdm}}, more than 40 (resp. 100) nodes are activated within the first 5 rounds of adaptive seeding. Also, within the first 10 rounds, all oracles successfully activated more than 80 users. Such observations suggest that there are likely several well followed nodes that once activated, has the potential to activate a big crowd. However, after the first 20 rounds, the margin between {\tt{grd\_lf}} and the rest of the baseline oracles ({\tt{bgg\_dgr}}, {\tt{rdm}}, {\tt{grd\_lnf}}) keeps increasing. From Figure \ref{fig:434433610}, we can see that both {\tt{grd\_kw}} and {\tt{grd\_lf}} were able to activate all 145 nodes in the subnetwork within 70 rounds in almost all 10 realizations. The remaining three oracles achieved only around 110.  From Figure \ref{fig:745823}, we can see that both {\tt{grd\_kw}} and {\tt{grd\_lf}} were able to activate almost all 243 nodes in the subnetwork within 120 rounds in almost all 10 realizations. The remaining three oracles achieved under 170.
	 
	\bibliographystyle{informs2014} 
	\bibliography{source.bib} 
	
		\begin{APPENDICES}
			
			\section{$\alpha,\beta$-Approximate Greedy Using RR Sets} \label{apd:RR}
			
			In Section \ref{sct:online}, we assume the existence of a $\alpha,\beta$-approximate oracle $\ora$. Taken as input a graph structure $\cG$, a node reward function $\br$, an influence probability function $w$ and an observed history $\hist{\by}$, $\ora(\cG, \br, w, \hist{\by})$ returns a node $s$ such that $\Delta(s|\ehist{\by}) \geq \frac{1}{\alpha} \max_{u\in \cG}\Delta(u|\ehist{\by})$ with probability at least $\beta$, where $\Delta(\cdot|\cdot)$ is calculated with respect to $\cG$, $w$, and $\br$. Here we present an example of $\ora$. 
			
			\subsection{Reverse Reachable (RR) Set}
			To precisely explain $\ora$, we first introduce the concept of \textit{reverse reachable sets} that was proposed by \cite{DBLP:journals/corr/abs-1212-0884}.
			\begin{definition}[\cite{Tang:2014:IMN:2588555.2593670}: Reverse Reachable Set]
				Let $v$ be a node in $\mathcal{V}$, and $\mathcal{H}$ be a graph obtained by removing each directed edge $e \in \cE$ with probability $1-w(e)$. 
				The reverse reachable (RR) set for $v$ in $\mathcal{H}$ is the set of nodes in $\mathcal{H}$ that can reach $v$. 
				That is, a node $u$ is in the RR set if and only if there is a directed path from $u$ to $v$ in $\mathcal{H}$.
			\end{definition}

			\cite{DBLP:journals/corr/abs-1212-0884} proved the following lemma that connects a RR Set for $v \in \cV$ with the probability $v$ is activated by a seed node $s \in \cV$:
			\begin{lemma}[\cite{DBLP:journals/corr/abs-1212-0884}]
				\label{lemma:cover} 
				For any seed $s \in \cV$ and node $v \in \cV$, the probability that a diffusion process from $s$ which follows the IC model with influence probabilities $w$ can activate $v$ equals the probability that $s$ is contained in an RR set for $v$ in a graph $\mathcal{H}$ generated by removing each directed edge $e$ in $\cV$ with probability $1-w(e)$.
			\end{lemma}
		
			To idea of our $\ora$ example is to sample RR sets to estimate $\Delta(s| \ehist{\by})$ for each $s \in \cV$ and output the $s$ with the biggest estimated value. Let $I$ be the set of nodes that have been activated during the observed history $\hist{\by}$, we defined a notion of \textit{random RR Set} that is similar to the one in \cite{Tang:2014:IMN:2588555.2593670} but different in the sense that only a subset of nodes $\cS \subseteq \cV$ are considerd:
			
			\begin{definition}[Random RR Set with respect to $\cS$]
				\label{def:RR-set}
				Let $\mathcal{W}$ be the distribution on $\mathcal{H}$ induced by the randomness in edge removals from $\mathcal{V}$. 
				A random RR set is an RR set generated on an instance of $\mathcal{H}$ randomly sampled from $\mathcal{W}$, for a node selected uniformly at random from $\cS$.
			\end{definition}
			
				\cite{Tang:2014:IMN:2588555.2593670} give an algorithm for generating a random RR set with respect to $\cS$ as follows. Let $w$ be the influence probabilities vector.

			\textbf{}
			
			\begin{breakablealgorithm}
				\caption{Random RR set}\label{alg:RRset}
				\begin{algorithmic}
					\State Initiate $R=\emptyset$
					\State Initiate an empty first-in-first-out queue $Q$
					\State Sample a node $v$ uniformly at random from $\mathcal{S}$, add to $R$
					\For{$u \in \mathcal{V}$ s.t. $(u, v) \in \cE$}
					\State Flip a biased coin with probability $w(u,v)$ of turning head. 
					\If{the coin turns head}
					\State Add $u$ to $Q$ and $R$
					\EndIf
					\EndFor
					\While{$Q$ is not empty}
					\State Extract the node $v'$ at the top of $Q$
					\For{$u' \in \mathcal{V}$ s.t. $(u', v') \in \cE$}
					\State Flip a biased coin with probability $w(u',v')$ of turning head. 
					\If{the coin turns head}
					\State Add $u'$ to $Q$ and $R$
					\EndIf
					\EndFor
					\EndWhile
				\end{algorithmic}
			\end{breakablealgorithm}
			\textbf{}
			
			From Lemma \ref{lemma:cover}, we can easily show the following result which connects the random RR sets sampling with the estimation of $\Delta(s|\ehist{\by})$:
			
			\begin{lemma} \label{lemma:expRR}
				Let $R_1, R_2, ..., R_m$ be i.i.d. random RR sets with respect to $\cV \setminus I$ and $v_i$ be the node of which $R_i$ is a RR set. For any $s \in \cV$, with $$\bar{\Delta}(s|\ehist{\by}) := \frac{\sum_{i = 1}^m \br(v_i)\mathbf{1}(s \in R_i)}{m/|\cV \setminus I|},$$
				we have
				\begin{equation}
				\E(\bar{\Delta}(s|\ehist{\by})) = \Delta(s|\ehist{\by}).
				\end{equation}
				\end{lemma}	
			
			\proof{Proof.}
			Let $\bX_i = \br(v_i)\mathbf{1}(s \in R_i) \cdot |\cV \setminus I|$. Then $\bX_i$'s are i.i.d. with mean 
			\begin{equation}
			\begin{split}
			\E(\bX_i) &= |\cV \setminus I| \cdot \sum_{v \in \cV \setminus I}\E(\br(v)\mathbf{1}(s \in R)|R\text{ is generated for }v) \cdot \frac{1}{|\cV \setminus I|}\\
			& = \sum_{v \in \cV \setminus I} \br(v) \pr(v\text{ is activated by }s) = \Delta(s|\ehist{\by}).
			\end{split}
			\end{equation}
			\endproof
			
			The plan is to sample enough such random RR sets with respect to $\cV \setminus I$ such that with high probability, the sample average $\bar{\Delta}(s|\ehist{\by})$ will be close enough to the true mean $\Delta(s|\ehist{\by})$. We now analyze how big $m$ needs to be for this to happen. For simplicity of presentation we temporarily denote $\Delta(s|\ehist{\by})$ and $\bar{\Delta}(s|\ehist{\by})$ with $\Delta(s)$ and $\bar{\Delta}(s)$ respectively. Since $\br(v_i)\mathbf{1}(s \in R_i) \cdot |\cV \setminus I| \in [0, |\cV \setminus I|]$ and the length of the interval is upper bounded by $n^2$ where $n := |\cV|$, with the Chernoff-Hoeffding concentration result, we have from Lemma \ref{lemma:expRR} that for any $s\in \cV$, 
			\begin{equation} \label{eq:55}
			\begin{split}
			&\pr(\bar{\Delta}(s) - \Delta(s) \geq x) \leq e^{-\frac{2mx^2}{n^2}},  \\
			&\pr(\Delta(s) - \bar{\Delta}(s) \geq x) \leq e^{-\frac{2mx^2}{n^2}} . 
			\end{split}
			\end{equation}
			
			Let $\alpha > 1, \beta > 0$ be given, set $m \geq \frac{2\alpha^2 n^2 \log(3/(1-\beta))}{(\alpha - 1)^2 \br^2_{\max}}$. Let $s^* \in \argmax_{s \in \cV} \Delta(s)$, and $s^{on} \in \argmax_{s \in \cV} \bar{\Delta}(s)$ we have the following:
			
			\begin{equation} \label{eq:56}
			\begin{split}
			&\pr\left(\Delta(s^{on}) \geq \frac{1}{\alpha} \Delta(s^*)\right)\\
			& = \pr \left(\Delta(s^*) - \Delta(s^{on}) \leq \frac{(\alpha - 1)}{\alpha} \Delta(s^*)\right)\\
			& = \pr\left(\Delta(s^*) - \bar{\Delta}(s^*) + \bar{\Delta}(s^*) - \bar{\Delta}(s^{on}) + \bar{\Delta}(s^{on}) - \Delta(s^{on}) \leq \frac{(\alpha - 1)}{\alpha} \Delta(s^*)\right)\\
			& \geq \pr\left(\Delta(s^*) - \bar{\Delta}(s^*) + \bar{\Delta}(s^{on}) - \Delta(s^{on}) \leq \frac{(\alpha - 1)}{\alpha} \Delta(s^*)\right)\\
			& \geq \pr\left(\Delta(s^*) - \bar{\Delta}(s^*) \leq \frac{(\alpha - 1)}{2\alpha} \Delta(s^*),  \bar{\Delta}(s^{on}) - \Delta(s^{on}) \leq \frac{(\alpha - 1)}{2\alpha} \Delta(s^*)\right),
			\end{split}
			\end{equation} 
			
			where the first inequality follows from the fact that $\bar{\Delta}(s^{*}) - \bar{\Delta}(s^{on}) \leq 0$.
			
			Now let $A = \{\Delta(s^*) - \bar{\Delta}(s^*) \leq \frac{(\alpha - 1)}{2\alpha} \Delta(s^*)\}, B = \{ \bar{\Delta}(s^{on}) - \Delta(s^{on}) \leq \frac{(\alpha - 1)}{2\alpha} \Delta(s^*)\}$. We have that $\pr(A, B) = 1 - \pr(A^c, B) - \pr(A, B^c) - \pr(A^c, B^c) \geq 1 - 2\pr(A^c) - \pr(B^c)$. From \eqref{eq:55}, we have that 
			\begin{equation} \label{eq:57}
			\pr(A^c), \pr(B^c) \leq e^{-2m\frac{(\alpha - 1)^2}{4\alpha^2 n^2} \Delta(s^*)^2} \leq e^{-2\frac{2\alpha^2 n^2 \log(3/(1-\beta))}{(\alpha - 1)^2 \br^2_{\max}} \frac{(\alpha - 1)^2}{4\alpha^2 n^2} \Delta(s^*)^2} \leq e^{-\log(3/(1-\beta))} = \frac{1-\beta}{3},
			\end{equation}
			where the second inequality follows from $m \geq \frac{2\alpha^2 n^2 \log(3/(1-\beta))}{(\alpha - 1)^2 \br^2_{\max}}$ and the last inequality follows from $\Delta(s^*) \geq \br_{\max}$ where $\br_{\max} = \max_{v\in \cV} \br(v)$.
			
			\eqref{eq:56} and \eqref{eq:57} toegther give us that
			
			\begin{equation}
			\begin{split}
			&\pr\left(\Delta(s^{on}) \geq \frac{1}{\alpha} \Delta(s^*)\right)\\
			&\geq \pr\left(\Delta(s^*) - \bar{\Delta}(s^*) \leq \frac{(\alpha - 1)}{2\alpha} \Delta(s^*),  \bar{\Delta}(s^{on}) - \Delta(s^{on}) \leq \frac{(\alpha - 1)}{2\alpha} \Delta(s^*)\right)\\
			& = \pr(A, B) \geq 1 - 2\pr(A^c) - \pr(B^c) \geq 1 - 3 \frac{1-\beta}{3} = \beta.
			\end{split}
			\end{equation}
			
			We summarize $\ora$ as below:
			
			\textbf{}
			
			\begin{breakablealgorithm}
				\caption{\ora}\label{alg:ora}
				\begin{algorithmic}
					\State \textbf{Input:} $\cG, I, \br, w, \hist{\by}, \alpha, \beta$
					\State Sample $m = \lceil \frac{2\alpha^2 n^2 \log(3/(1-\beta))}{(\alpha - 1)^2 \br^2_{\max}} \rceil$ random RR sets $R_1, ..., R_m$ with respect to $\cV \setminus I$ on graph $\cG$ with influence probabilities $w$
					\For{$s \in \cV$}
					\State Compute $\bar{\Delta}(s|\ehist{\by}) := \frac{\sum_{i = 1}^m \br(v_i)\mathbf{1}(s \in R_i)}{m/|\cV \setminus I|}$
					\EndFor
					\State Output $s^{on} \in \argmax_{s \in \cV}\bar{\Delta}(s|\ehist{\by})$
				\end{algorithmic}
			\end{breakablealgorithm}
		
		\textbf{}
		
		From the proceeding analysis, we know that $\ora$ is a $\alpha, \beta$-approximate greedy algorithm.
			\end{APPENDICES}
\end{document}